\documentclass[preprint,12pt,times]{elsarticle}

\usepackage{amsmath,amsfonts}
\usepackage{algorithmic}
\usepackage{algorithm}
\usepackage{array}
\usepackage[caption=false,font=normalsize,labelfont=sf,textfont=sf]{subfig}
\usepackage{textcomp}
\usepackage{stfloats}
\usepackage{url}
\usepackage{verbatim}
\usepackage{graphicx}
\usepackage{enumitem}
\usepackage{xcolor} 
\usepackage[hidelinks]{hyperref}
\hypersetup{
  pdftitle={Leakage-Free Evaluation and Distribution-Robust Spatio-Temporal Graph Learning for Inductive Kriging},
  pdfauthor={Chen Yang, Changhao Zhao, Haoyang Zhao, Youquan He, Chen Wang, and Jiansheng Fan},
  pdfsubject={Accepted manuscript, Pattern Recognition, DOI: 10.1016/j.patcog.2026.114638},
  pdfkeywords={spatio-temporal learning, inductive kriging, graph neural networks, out-of-distribution generalization, data leakage}
}
\usepackage{caption}
\usepackage{multirow}
\usepackage{booktabs}
\usepackage{tabularx}
\usepackage{amssymb}
\usepackage{bm}

\journal{Pattern Recognition}

\begin{document}

\begin{frontmatter}

\title{Leakage-Free Evaluation and Distribution-Robust Spatio-Temporal Graph Learning for Inductive Kriging\tnoteref{accepted-manuscript}}
\tnotetext[accepted-manuscript]{Accepted manuscript. The Version of Record is available in \textit{Pattern Recognition} at \href{https://doi.org/10.1016/j.patcog.2026.114638}{doi:10.1016/j.patcog.2026.114638}. \textcopyright{} 2026. This manuscript version is made available under the \href{https://creativecommons.org/licenses/by-nc-nd/4.0/}{CC BY-NC-ND 4.0 license}.}

\author[1]{Chen Yang\fnref{fn1}}
\author[2]{Changhao Zhao\fnref{fn1}}
\author[1]{Haoyang Zhao}
\author[1]{Youquan He}

\author[1]{Chen Wang\corref{cor1}}
\ead{chwang@tsinghua.edu.cn}

\author[1]{Jiansheng Fan}

\cortext[cor1]{Corresponding author.}
\fntext[fn1]{These authors contributed equally to this work.}

\affiliation[1]{organization={Department of Civil Engineering, Tsinghua University},
                city={Beijing},
                postcode={100084},
                country={China}}

\affiliation[2]{organization={School of Computer and Cyber Sciences, Communication University of China},
                city={Beijing},
                postcode={100024},
                country={China}}

\begin{abstract}
Inductive kriging estimates values at unobserved locations from sparse sensor data, enabling continuous field reconstruction when dense deployment is impractical. However, common 2$\times$2 and 2$\times$3 evaluation protocols can leak spatial information through model selection and obscure true out-of-distribution (OOD) behavior. We propose a leakage-free 3$\times$3 partition that separates training, validation, and testing in both space and time, so that model fitting, checkpoint selection, and final reporting are performed on distinct spatio-temporal domains. Under this stricter setting, we introduce DRIK (Distribution-Robust Inductive Kriging), a framework with three task-specific mechanisms: Spatial Continuity Regularization (SCR) perturbs coordinates to reduce dependence on one discretized graph; Masked Flow Disambiguation (MFD) prunes ambiguous propagation from zero-padded masked nodes; and Structural Domain Expansion (SDE) uses validation-node topology without labels to reduce train-inference structural mismatch. Experiments on six spatio-temporal datasets show that DRIK consistently outperforms state-of-the-art baselines, reducing MAE by up to 12.48\% and achieving lower test-to-validation MAE ratios under leakage-free evaluation. These results indicate that robust inductive kriging requires both leakage-free evaluation and mechanisms that explicitly address the structural shifts introduced by unseen nodes.
\end{abstract}


\begin{keyword}
spatio-temporal learning \sep inductive kriging \sep graph neural networks \sep out-of-distribution generalization \sep data leakage
\end{keyword}

\end{frontmatter}

\section{Introduction}

Sensor networks are widely used to monitor dynamic systems such as traffic flow \cite{ref1}, air quality \cite{ref2}, and solar energy production \cite{ref3}. Because dense deployment is costly and sometimes physically infeasible, many locations remain unsensed, limiting spatial coverage over large areas \cite{ref4,ref5}. This sparsity is especially problematic when downstream decisions require fine-grained maps rather than measurements only at installed stations. Inductive kriging addresses this issue by estimating values at unseen locations from existing sensors, enabling high-resolution field reconstruction with fewer deployed nodes \cite{ref6,ref7,ref8}. In contrast to standard missing-value imputation on a fixed graph, the target locations in inductive kriging are absent during model development, which makes spatial extrapolation and graph generalization central to the task.

\begin{figure}[t]
\centering
\includegraphics[width=\linewidth]{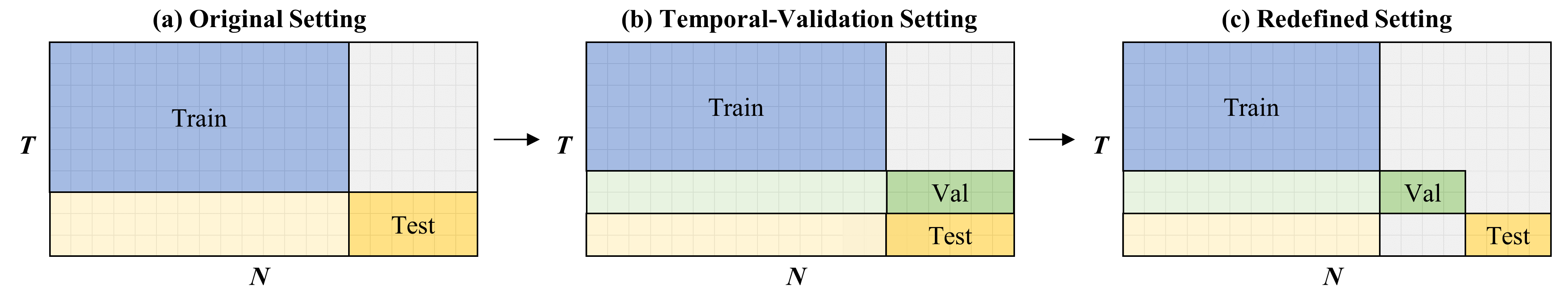}
\caption{Comparison of inductive kriging settings. For better visualization, the matrix has been transposed. Blue, green, and yellow indicate the training, validation, and test sets, respectively. Light green and light yellow represent observed data from the validation and test periods used for prediction, while gray denotes data that remain unused throughout the process.}
\label{fig_1}
\vspace{-10pt}
\end{figure}

The conventional protocol for deep learning-based inductive kriging \cite{ref6} follows the 2$\times$2 split in Fig.~\ref{fig_1}(a). The dataset $\bm{X}\in {{\mathbb{R}}^{N\times T}}$ is divided along temporal and spatial dimensions, with training and test sets placed on diagonal blocks. During training, the model reconstructs masked nodes from observed nodes; during testing, it predicts test nodes using training nodes from the test period. This protocol leaks information when early stopping or model selection relies on test loss \cite{ref7}, because the supposedly unseen test block participates in the optimization loop through checkpoint selection. The 2$\times$3 variant adds a validation period (Fig.~\ref{fig_1}(b)) \cite{ref8,ref10}, but validation and test data still share the same spatial nodes. As a result, the validation procedure can repeatedly expose the model-selection process to the spatial distribution used for final evaluation, leaving spatial leakage unresolved even when temporal leakage is reduced.

We therefore adopt a leakage-free 3$\times$3 partition, as shown in Fig.~\ref{fig_1}(c). Training, validation, and test sets are separated in both space and time, so model fitting, model selection, and final evaluation are performed on distinct spatio-temporal domains. This setting exposes the out-of-distribution (OOD) nature of inductive kriging \cite{ref11,ref12}: prediction at unseen nodes changes graph density, neighborhood composition, and propagation paths, creating a structural shift that fixed-topology GNNs are not designed to handle \cite{ref8}.

To address these shifts, we propose DRIK (Distribution-Robust Inductive Kriging), a framework tailored to leakage-free inductive kriging. The design follows the observation that the inductive setting creates three coupled sources of instability: the coordinate graph is only a discretized approximation of a continuous field, masked target nodes initially carry no reliable values, and the graph encountered at inference time is denser than the graph used for supervised training. DRIK therefore combines three mechanisms: Spatial Continuity Regularization (SCR) perturbs node coordinates to reduce dependence on a single discretized graph; Masked Flow Disambiguation (MFD) prunes ambiguous message flow from zero-padded masked nodes; and Structural Domain Expansion (SDE) incorporates validation-node topology without labels to reduce the structural gap between training and inference. Our contributions are summarized as follows:
\begin{itemize}[leftmargin=1.5em, itemsep=0.5ex]
    \item \textbf{Redefinition of Protocol:} We redefine inductive kriging with a leakage-free 3$\times$3 partition that removes information leakage and aligns evaluation with realistic deployment settings.
    
    \item \textbf{Methodological Innovation:} We identify distribution shift as a bottleneck for inductive kriging and propose DRIK, which employs a three-tier strategy (SCR, MFD, SDE) to explicitly learn continuous, disambiguated, and domain-expanded representations for enhanced OOD robustness.
    
    \item \textbf{Empirical Results:} Extensive experiments on six datasets demonstrate that DRIK consistently outperforms existing methods, reducing error by up to 12.48\% and exhibiting stronger generalization, evidenced by a lower test-to-validation MAE ratio.
\end{itemize}

\section{Related work}
This section reviews inductive kriging and graph OOD generalization, focusing on the evaluation leakage and distribution-shift issues that motivate DRIK.

\subsection{Inductive kriging}
Spatio-temporal kriging approaches are generally categorized into transductive and inductive paradigms. Transductive kriging treats the task as data imputation within a fixed set of locations known during training. Existing methods typically employ matrix or tensor factorization on static tensors \cite{ref13,ref14,ref15,ref16}, combine Graph Neural Networks (GNNs) with recurrent architectures (e.g., GRIN \cite{ref17}, GNN-ODFill \cite{ref18}, and GinAR+ \cite{ref45}) to capture dependencies, or utilize generative probabilistic frameworks like PriSTI \cite{ref20} to reconstruct missing entries. These methods improve prediction or imputation under fixed sensor sets, but they still rely on predefined graph structures and cannot directly generalize to unseen locations without retraining \cite{ref21}, necessitating the use of inductive kriging for scalable sensing in dynamic environments.

Inductive kriging estimates values for sensors absent from training. KCN \cite{ref22} introduced localized inductive learning with $k$-nearest-neighbor graphs, while IGNNK \cite{ref6} trained diffusion graph convolutions on randomly sampled subgraphs so that the model can reconstruct signals from partial observations. Later methods improved the spatio-temporal encoder: SATCN \cite{ref23} combines temporal convolutions with sparsity-aware masking, LSJSTN \cite{ref24} separates long-term static trends from short-term fluctuations, INCREASE \cite{ref7} incorporates heterogeneous urban relations, and IAGCN \cite{ref25} learns adaptive graph correlations. KITS \cite{ref8} further targets the train-inference graph gap by inserting virtual nodes during training. These advances cover local proximity modeling, subgraph reconstruction, temporal decomposition, relation-aware representation learning, and graph-gap simulation. Nevertheless, they still largely operate under evaluation protocols where model selection may reuse the spatial domain of final testing, making it difficult to isolate true spatial extrapolation from adaptation to a repeatedly accessed node topology.

This protocol issue is central to our study. Common 2$\times$2 or 2$\times$3 splits can expose the model-selection process to the same spatial nodes used for final testing, causing implicit spatial leakage. As a result, reported performance may overestimate generalization to truly unseen regions and understate the OOD difficulty of inductive kriging.

\subsection{OOD generalization on graphs}
OOD generalization on graphs is challenging because real-world graph data often shift across environments, sampling processes, and operating conditions \cite{ref11,ref12,ref26}. Similar robustness concerns also arise in broader deep learning applications for dynamic engineering systems, such as battery charging optimization and remaining-useful-life prediction \cite{ref46,ref47}. In graph learning, representative solutions include disentanglement methods such as DisenGCN \cite{ref27} and FactorGCN \cite{ref28}, causal methods such as OOD-GNN \cite{ref29} and CAM \cite{ref30}, and optimization strategies such as DIR \cite{ref31} and WT-AWP \cite{ref32}.

Graph augmentation is another major robustness strategy. Node-level methods such as GRAND \cite{ref33} and PUA \cite{ref34} perturb attributes or pseudo-features; edge-level methods such as GAug \cite{ref35} and MH-Aug \cite{ref36} vary graph connectivity; and subgraph-level methods such as GraphCL \cite{ref37} and GREA \cite{ref38} construct contrastive structural views. These approaches improve invariance by simulating perturbations during training.

However, inductive kriging has task-specific OOD factors that generic augmentation does not directly capture: continuous spatial coordinates, zero-padded masked nodes, and validation-node topology that can be used without labels. These factors are not merely implementation details. Coordinates determine how the graph is discretized, masked nodes determine whether message passing starts from meaningful values or artificial zeros, and validation topology provides unlabeled structural information about a spatial domain that is disjoint from supervised training nodes. DRIK is designed around these properties rather than applying generic graph perturbations alone.

\section{Methodology}
\subsection{Leakage-free inductive kriging formulation}
\subsubsection{Task definition and notation}
Inductive kriging infers values at locations absent during training from sparse observations collected by deployed sensors. Unlike transductive imputation on a fixed node set, it requires extrapolation from observed sensor locations to entirely unobserved ones, which is common when dense deployment is infeasible.

Let ${{\mathcal{V}}_{\text{o}}}=\{{{v}_{1}},\ldots ,{{v}_{{{N}_{\text{o}}}}}\}$ denote the set of observed nodes and ${{\mathcal{V}}_{\text{u}}}=\{{{u}_{1}},\ldots ,{{u}_{{{N}_{\text{u}}}}}\}$ denote the set of unobserved nodes, where ${{N}_{\text{o}}}$ and ${{N}_{\text{u}}}$ are the numbers of observed and unobserved locations, respectively. Each node is associated with spatial coordinates ${{s}_{v}}\in {{\mathbb{R}}^{d}}$, where $d$ is typically 2. For the observed nodes, historical measurements over the most recent $t$ time steps are denoted by $\bm{X}_{\text{o}}^{T-t:T}\in {{\mathbb{R}}^{{{N}_{\text{o}}}\times t}}$, where $T$ is the current inference time and $t$ is the temporal input window. The goal of inductive kriging is to estimate the unknown values on ${{\mathcal{V}}_{\text{u}}}$ at time $T$ using the historical observations on ${{\mathcal{V}}_{\text{o}}}$ together with the spatial relations among nodes.

Following prior graph-based inductive kriging studies \cite{ref6,ref8}, we construct a spatial graph from node coordinates using a Gaussian kernel \cite{ref39} over pairwise distances. When only observed nodes are considered, this yields a training graph ${{\mathcal{G}}_{\text{o}}}=({{\mathcal{V}}_{\text{o}}},{\bm{A}_{\text{o}}})$, where ${\bm{A}_{\text{o}}}\in {{[0,1]}^{{{N}_{\text{o}}}\times {{N}_{\text{o}}}}}$ is the adjacency matrix over observed nodes. During inference, kriging is performed on the enlarged graph $\mathcal{G}=({{\mathcal{V}}_{\text{o}}}\cup {{\mathcal{V}}_{\text{u}}},\bm{A})$, whose adjacency matrix can be partitioned as:
\begin{equation}
\bm{A}=\left[ \begin{matrix}
   {\bm{A}_{\text{oo}}} & {\bm{A}_{\text{ou}}}  \\
   {\bm{A}_{\text{uo}}} & {\bm{A}_{\text{uu}}}  \\
\end{matrix} \right],\quad {\bm{A}_{\text{oo}}}={\bm{A}_{\text{o}}}
\end{equation}
where ${\bm{A}_{\text{oo}}}$ models relations among observed nodes, ${\bm{A}_{\text{ou}}}$ and ${\bm{A}_{\text{uo}}}$ model cross-relations between observed and unobserved nodes, and ${\bm{A}_{\text{uu}}}$ models relations among unobserved nodes. The task is therefore not merely to recover missing values on a fixed graph, but to extrapolate from the observed-node domain ${{\mathcal{V}}_{\text{o}}}$ to the unseen-node domain ${{\mathcal{V}}_{\text{u}}}$.

To ensure a rigorous evaluation, we adopt a leakage-free 3$\times$3 spatio-temporal partition, where training, validation, and test sets are separated in both space and time. Model fitting is performed only on the training split, model selection relies only on the validation split, and final performance is reported only on the test split. Under this strict setting, inductive kriging naturally becomes an out-of-distribution problem. Although the partition is disjoint in both dimensions, the task-specific challenge mainly arises from the spatial domain, since the defining property of inductive kriging is prediction at previously unseen locations, which directly changes node neighborhoods, graph connectivity, and propagation patterns.

The purpose of the 3$\times$3 partition is therefore not only to reserve an additional validation block, but to make model selection and final testing occur on different spatial domains. In conventional 2$\times$2 or 2$\times$3 protocols, the validation or early-stopping procedure can still expose the model selection process to the same node domain used for final evaluation. The proposed split prevents this spatial-domain reuse, so the reported test performance reflects extrapolation to an independent set of unseen locations rather than adaptation to a repeatedly accessed sensor topology.

\subsubsection{OOD risk formulation}
Under the above setting, training and evaluation are conducted on different graph domains and correspond to different learning objectives. During training, the model only has access to the observed graph ${{\mathcal{G}}_{\text{o}}}=({{\mathcal{V}}_{\text{o}}},{\bm{A}_{\text{oo}}})$. A common self-supervised strategy is to randomly mask a subset of observed nodes and reconstruct their values from the remaining observations. Let ${{f}_{\theta }}$ denote the kriging model parameterized by $\theta $, and let $\ell (\cdot ,\cdot )$ be the loss function. The empirical risk minimized during training can then be written as:
\begin{equation}
{{\mathcal{R}}_{\text{tr}}}(\theta )={{\mathbb{E}}_{({\bm{X}_{\text{o}}},{\bm{A}_{\text{oo}}}) \sim {{P}_{\text{o}}}}}[\ell ({{f}_{\theta }}({\bm{X}_{\text{o}}};{\bm{A}_{\text{oo}}}),{\bm{Y}_{\text{o}}})]
\end{equation}
where ${{P}_{\text{o}}}$ denotes the distribution over observed-node subgraphs and ${\bm{Y}_{\text{o}}}$ denotes the masked reconstruction targets on observed nodes.

During validation and testing, however, the model is evaluated on the enlarged graph $\mathcal{G}=({{\mathcal{V}}_{\text{o}}}\cup {{\mathcal{V}}_{\text{u}}},\bm{A})$, where previously unseen nodes are inserted into the graph and their values must be inferred from observed measurements. The corresponding evaluation risk is:
\begin{equation}
{{\mathcal{R}}_{\text{eval}}}(\theta )={{\mathbb{E}}_{({\bm{X}_{\text{o}}},\bm{A}) \sim {{P}_{\text{ou}}}}}[\ell ({{f}_{\theta }}({\bm{X}_{\text{o}}};\bm{A}),{\bm{Y}_{\text{u}}})]
\end{equation}
where ${{P}_{\text{ou}}}$ denotes the distribution over enlarged graphs containing both observed and unseen nodes, and ${\bm{Y}_{\text{u}}}$ denotes the prediction targets on unobserved nodes.

The discrepancy between ${{\mathcal{R}}_{\text{tr}}}$ and ${{\mathcal{R}}_{\text{eval}}}$ arises from both the target domain and the graph operator. Training optimizes masked reconstruction on observed nodes, whereas evaluation measures extrapolation to unseen nodes. Meanwhile, the propagation operator changes from the training-time operator ${\bm{A}_{\text{oo}}}$ to the inference-time operator $\bm{A}$. Since the insertion of unseen nodes changes node degrees, neighborhood composition, and graph spectrum, the hidden representations produced by graph propagation are subject to coupled structural and covariate shifts.

This OOD nature can also be understood from the perspective of classical kriging. Let $\bm{K}$ denote the kernel matrix induced by the spatial covariance function. The conditional mean at unseen nodes is given by
\begin{equation}
{\bm{\mu }_{\text{u}|\text{o}}}={\bm{K}_{\text{uo}}}\bm{K}_{\text{oo}}^{-1}{\bm{X}_{\text{o}}}
\end{equation}
where ${\bm{K}_{\text{oo}}}$ captures covariance within the observed domain and ${\bm{K}_{\text{uo}}}$ captures covariance between unseen and observed nodes. Self-supervised training on ${{\mathcal{V}}_{\text{o}}}$ primarily learns within-domain reconstruction, whereas kriging at evaluation time additionally depends on cross-domain extrapolation through ${\bm{K}_{\text{uo}}}$. This mismatch makes leakage-free inductive kriging a natural spatial OOD learning problem.

This formulation provides the analytical motivation for evaluating validation and test performance on spatially disjoint domains: a model that performs well on within-domain reconstruction or on a reused validation topology may still fail when the cross-domain covariance structure and graph operator change at test time.

\begin{figure}[t]
\centering
\includegraphics[width=\linewidth]{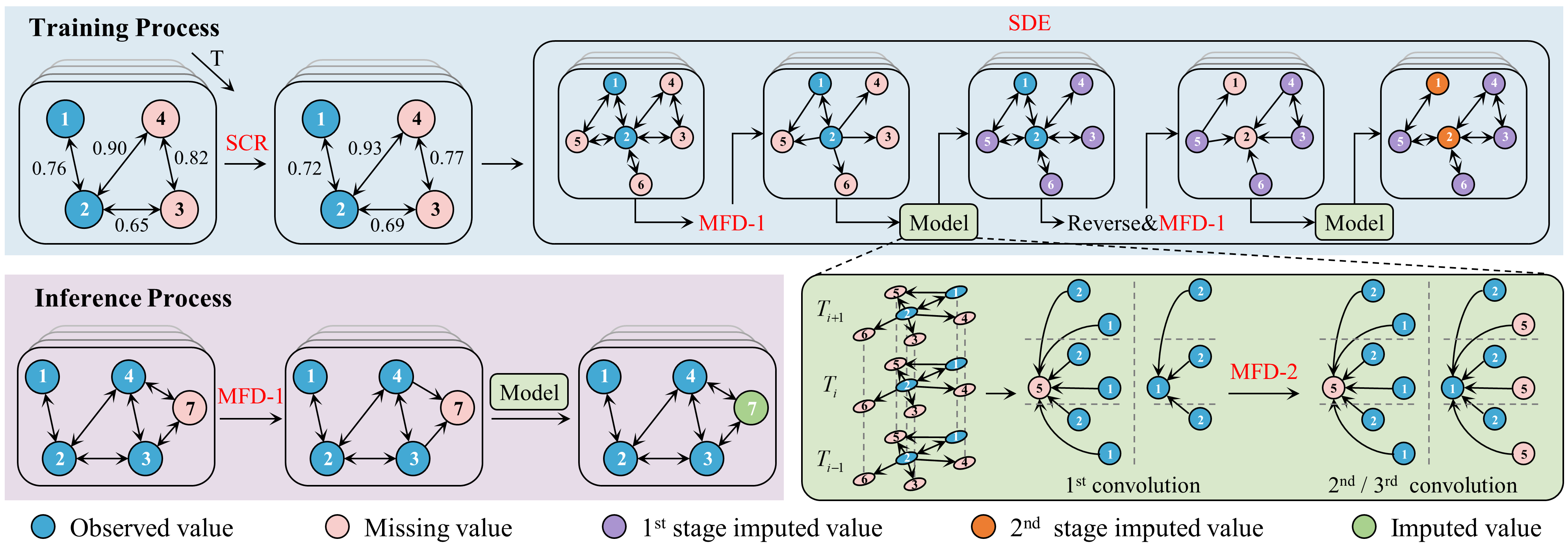}
\caption{Overview of DRIK. Spatial Continuity Regularization (SCR) and Structural Domain Expansion (SDE) are employed exclusively during the training process. Masked Flow Disambiguation (MFD) is applied in both the training and inference processes to ensure structural consistency. The inset details the spatio-temporal convolution mechanism of the kriging model across temporal intervals and spatial neighbors; taking node-1 (observed) and node-5 (masked) as examples, it illustrates how MFD-2 reinstates the outgoing edges from the masked node to the observed node in subsequent convolutions to safely restore graph connectivity.}
\label{fig_2}
\vspace{-10pt}
\end{figure} 

\subsection{Overview of DRIK}
We propose DRIK to address the structural and covariate shifts exposed by the leakage-free inductive setting. Instead of relying on a standalone decrement operation \cite{ref6}, a standalone increment operation \cite{ref8}, or generic graph augmentation, DRIK combines decrement and increment operations around three task-specific mechanisms:

\begin{itemize}[leftmargin=1.5em, itemsep=0.5ex]
    \item \textbf{Spatial Continuity Regularization (SCR).} SCR perturbs node coordinates during training, rebuilding local edge weights and reducing dependence on a single discretized graph.
    
    \item \textbf{Masked Flow Disambiguation (MFD).} It prunes ambiguous paths from masked nodes initialized with zeros, preventing unreliable signals from being treated as valid observations.
    
    \item \textbf{Structural Domain Expansion (SDE).} SDE inserts validation-node topology into the training graph without validation labels, reducing the density and topology gap between training and inference.
\end{itemize}

DRIK therefore differs from existing inductive kriging frameworks in both objective and mechanism. IGNNK focuses on transferable reconstruction over sampled subgraphs, INCREASE enriches heterogeneous spatio-temporal relations, and KITS narrows the graph gap with artificial virtual nodes. DRIK instead targets leakage-free structural robustness by jointly using local coordinate perturbation, task-aware edge pruning, and real validation-node topology without labels. This combination is important because no single operation fully resolves the leakage-free setting: perturbing coordinates improves invariance but does not remove zero-padding noise, pruning noisy edges stabilizes propagation but does not address graph-density mismatch, and validation topology reduces structural shift only when its unlabeled nodes are prevented from injecting unreliable messages.

Fig.~\ref{fig_2} summarizes the training pipeline. DRIK masks a subset of training nodes for self-supervised reconstruction, applies SCR to build a perturbed adjacency matrix, inserts fully masked validation nodes through SDE, and uses MFD-1 before propagation. The core kriging model first imputes the masked nodes, then a reverse pass masks the originally observed nodes and uses the imputed values as inputs. This two-stage routine trains the model on imperfect expanded graphs while keeping validation measurements hidden. At inference, test nodes are inserted into the observed graph and processed with the same structural rules.

Following KITS \cite{ref8}, DRIK uses a spatio-temporal kriging backbone that combines temporal aggregation with graph convolution. Let ${\bm{Z}^{(l)}}$ denote the hidden representation at layer $l$ and $\bm{Z}_{T-t:T}^{(l)}$ the features over the input window. To reduce trivial self-feature memorization and narrow the gap between observed and zero-padded nodes, STGC removes self-loops through the diagonal-masked adjacency ${\bm{A}^{-}}$:
\begin{equation}
{\bm{Z}^{(l+1)}}=\text{FC}(\text{GC}(\bm{Z}_{T-t:T}^{(l)},{\bm{A}^{-}}))
\end{equation}
where $\text{GC}(\cdot )$ is the inductive graph convolution and $\text{FC}(\cdot )$ is a fully connected layer. DRIK also uses Reference-based Feature Fusion (RFF) to align each unobserved node ${{u}_{i}}\in {{\mathcal{V}}_{\text{u}}}$ with its most similar observed node according to cosine similarity, with maximum score $\bm{S}_{i}^{*}$ and index $\bm{Ind}_{i}^{*}$:
\begin{equation}
{\bm{Z}_{{{u}_{i}}}}=\text{FC}({\bm{Z}_{{{u}_{i}}}}\parallel \bm{S}_{i}^{*}\odot \text{Align}({{\mathcal{V}}_{o}},\bm{Ind}_{i}^{*}))
\end{equation}
where $\parallel $ denotes concatenation, $\odot $ is element-wise multiplication, and $\text{Align}(\cdot )$ extracts the paired observed-node feature.

DRIK further manages graph connectivity across layers through MFD. MFD-1 first removes outgoing edges from masked nodes to block unreliable zero-padded signals. After one STGC layer, masked nodes have received information from observed neighbors, so MFD-2 restores their outgoing edges to observed nodes while keeping masked-to-masked edges pruned. This progressive relaxation restores connectivity without allowing early noisy features to circulate.

\subsection{Three-tier distribution-robust learning strategy}

We next detail the three robust-learning components of DRIK: SCR addresses coordinate-level graph discretization, MFD controls edge-level ambiguity from masked nodes, and SDE reduces subgraph-level density shifts by using validation-node topology without labels.

\subsubsection{Spatial Continuity Regularization}
Inductive kriging discretizes a continuous spatial process into a graph built from node coordinates and distances. When new nodes are inserted at inference time, local neighborhoods change, edge weights are recomputed, and the receptive field of each target can differ from the one observed during supervised training. SCR reduces this dependence on one fixed discretization by perturbing node locations within a local geometric boundary during training. The purpose is not to create arbitrary spatial noise, but to expose the kriging backbone to nearby graph realizations that are plausible under the same spatial field.

\begin{figure}[t]
\centering
\includegraphics[width=\linewidth]{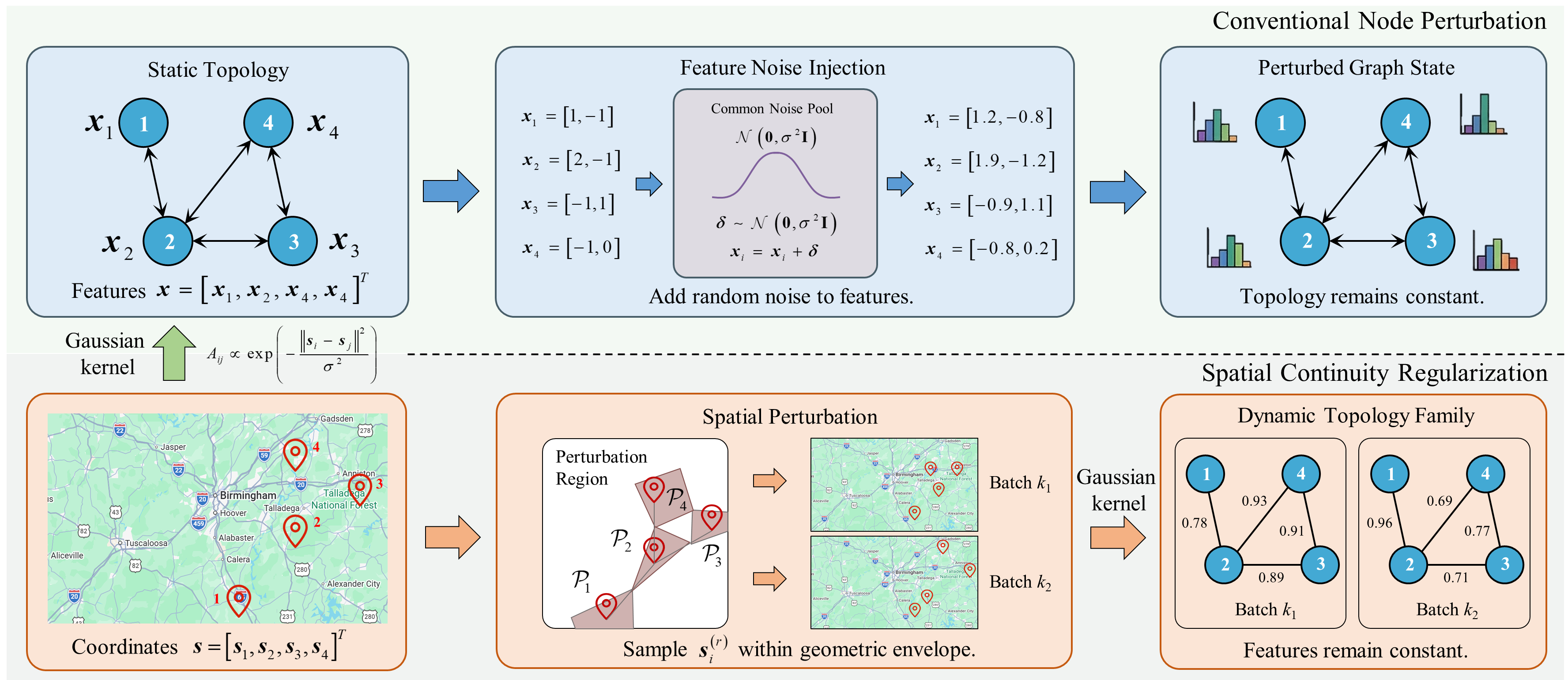}
\caption{Comparison between conventional node perturbation and Spatial Continuity Regularization (SCR). Conventional methods (top) inject noise into node features while maintaining a static graph topology. In contrast, SCR (bottom) perturbs spatial coordinates within a geometric envelope, creating dynamic graph topologies during training. This forces the model to learn coordinate-invariant representations and prevents overfitting to a fixed graph structure.}
\label{fig_3}
\vspace{-10pt}
\end{figure} 

Unlike node feature perturbation \cite{ref33,ref34}, SCR perturbs spatial coordinates and rebuilds the adjacency matrix at each training step (Fig.~\ref{fig_3}). The model is therefore exposed to local topological variations instead of only feature noise.

Let $\mathcal{V}$ be the set of training nodes, where each node $v\in \mathcal{V}$ is associated with spatial coordinates ${{s}_{v}}\in {{\mathbb{R}}^{d}}$ (typically $d=2$). Let $\mathcal{N}(v)$ denote the neighbor set of $v$ under the initial inductive graph. We define the continuous perturbation region ${{\mathcal{P}}_{v}}\subset {{\mathbb{R}}^{d}}$ as the convex hull of the midpoints between $v$ and its neighbors:
\begin{equation}
{{\mathcal{P}}_{v}}=\text{conv}\left( \left\{ {{m}_{v,u}}={{s}_{v}}+\frac{1}{2}({{s}_{u}}-{{s}_{v}})|u\in \mathcal{N}(v) \right\} \right)
\end{equation}
where the scale of ${{\mathcal{P}}_{v}}$ is intrinsically bounded by the local inter-node distances. The vertices of ${{\mathcal{P}}_{v}}$ are the midpoints between $v$ and its neighbors, and their convex hull forms a safe spatial envelope that preserves the broader global topology.

During each training iteration $r$, the spatial coordinates of all nodes are perturbed by uniformly sampling from their respective perturbation regions:
\begin{equation}
\tilde{s}_{v}^{(r)} \sim \text{Unif}({{\mathcal{P}}_{v}})
\end{equation}
	 
Following this perturbation, the adjacency matrix is dynamically rebuilt using a kernelized, row-normalized $k$-nearest-neighbor graph. The edge weight between node $v$ and $u$ is calculated as:
\begin{equation}
{{\tilde{\bm{A}}}^{(r)}}(v,u) = \frac{\exp \left( -\|\tilde{s}_{v}^{(r)}-\tilde{s}_{u}^{(r)}\|^{2} / \sigma^{2} \right)}{\sum\limits_{u'\in \text{kNN}_{k}(v)} \exp \left( -\|\tilde{s}_{v}^{(r)}-\tilde{s}_{u'}^{(r)}\|^{2} / \sigma^{2} \right)} \cdot \mathbb{I}\{u\in \text{kNN}_{k}(v)\}
\end{equation}
where $\sigma >0$ is the Gaussian kernel bandwidth that controls the decay rate of edge weights, and $\mathbb{I}\{\cdot \}$ is the indicator function. This continuity-aware perturbation exposes the kriging model to a continuous family of propagation operators $\{{{\tilde{\bm{A}}}^{(r)}}\}$, reducing its sensitivity to graph discretization.

For traffic datasets, SCR should be interpreted as local regularization of a coordinate graph approximation, not as an assumption that sensors physically move or that traffic propagation is purely Euclidean. Its perturbations are restricted to the convex hull of midpoint locations between a node and its neighbors, so they alter local edge weights without changing the global sensor layout. If detailed road network graphs or travel distance matrices are available, topology-constrained perturbations would be a natural extension.

\subsubsection{Masked Flow Disambiguation}
Inductive kriging must propagate over masked nodes that serve as self-supervised targets during training or represent unseen sensors during inference. Because these nodes are initialized with zeros, letting them send messages immediately can inject unreliable signals into observed neighbors and cause masked nodes to reinforce one another before receiving useful context. This issue is specific to inductive kriging: the absence of a value is not a noisy measurement but a structural condition of the target node. MFD addresses this with layer-wise edge pruning.

Unlike random edge dropping \cite{ref35,ref36}, MFD is deterministic, task-aware, and applied in both training and inference. Edge participation depends on whether a node is masked and on the convolution layer, which keeps structural rules consistent across domains.

Let $\mathcal{M}\subset \mathcal{V}$ be the set of masked nodes in the current graph, and let $\bm{A}$ represent the corresponding adjacency matrix. We define the layer-wise edge-drop operator ${{\Phi }^{(l)}}$, applied just before the $l$-th graph convolution layer. Before the first convolution ($l=0$), the first phase, MFD-1, removes all outgoing edges from the masked nodes:
\begin{equation}
{{\Phi }^{(0)}}{{(\bm{A})}_{vu}}=\bm{A}(v,u)\cdot \mathbb{I}\{v\notin \mathcal{M}\}
\end{equation}

This initial pruning prevents zero-padded masked nodes from contaminating unmasked neighbors or reinforcing one another before they have received reliable context.

For subsequent convolutions ($l\ge 1$), the dropping rule transitions to MFD-2, which is relaxed to selectively remove only the edges directly between masked nodes:
\begin{equation}
{{\Phi }^{(l)}}{{(\bm{A})}_{vu}}=\bm{A}(v,u)\cdot \mathbb{I}\{\neg (v\in \mathcal{M}\wedge u\in \mathcal{M})\}
\end{equation}
	 
After the first layer, masked nodes have aggregated context from unmasked neighbors, so their outgoing edges to unmasked nodes can be restored. Masked-to-masked edges remain pruned to avoid circular propagation of inaccurate information.

\subsubsection{Structural Domain Expansion}
The 3$\times$3 partition exposes a structural shift: graph density and neighborhood relations change when unseen nodes are inserted at inference time. A model trained only on the observed-node graph can therefore encounter different propagation paths during validation or testing even if the underlying physical process is unchanged. SDE reduces this gap by incorporating validation-node topology into training while strictly hiding validation measurements. In this way, the model can adapt to a richer spatial structure without using validation labels as supervised targets.

\begin{figure}[t]
\centering
\includegraphics[width=0.6\linewidth]{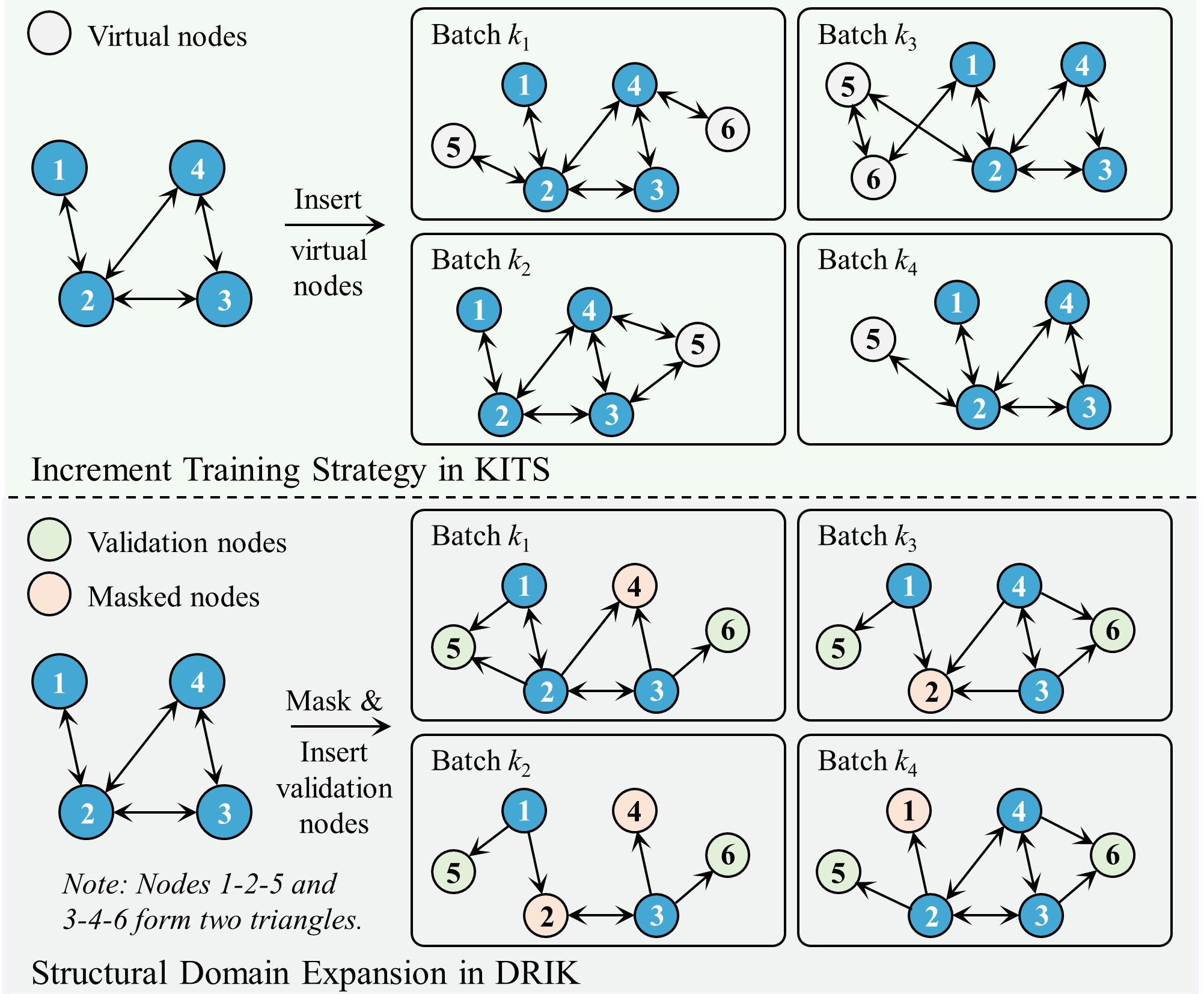}
\caption{Comparison of graph expansion operations between KITS and Structural Domain Expansion (SDE). Unlike KITS (top) which relies solely on an incremental approach with random virtual nodes, SDE (bottom) utilizes a combined decrement-increment strategy. SDE inserts actual validation nodes and applies MFD-1 to dynamically drop the outgoing edges of nodes without values per batch, preserving real-world spatial distribution while enhancing topological diversity.}
\label{fig_4}
\vspace{-10pt}
\end{figure} 

Fig.~\ref{fig_4} contrasts SDE with the increment strategy in KITS \cite{ref8}. KITS inserts random virtual nodes, which may increase graph density but can deviate from the real sensor distribution. SDE instead inserts actual validation nodes as unlabeled topology and combines this expansion with dynamic masking through MFD-1, providing realistic structural diversity without exposing validation targets.

Let ${{\mathcal{V}}_{\text{tr}}}$ and ${{\mathcal{V}}_{\text{val}}}$ denote disjoint training and validation node sets. SDE constructs an augmented graph ${{\mathcal{G}}_{\cup }}=({{\mathcal{V}}_\text{tr}}\cup {{\mathcal{V}}_\text{val}},{\bm{A}_{\cup }})$ and trains on it through a two-stage pseudo-labeling routine.

In the first pass, SDE masks a training subset ${{\mathcal{M}}_{1}}\subset {{\mathcal{V}}_{\text{tr}}}$ and all validation nodes. The model predicts both groups from available training measurements $\bm{X}$, and validation predictions are detached as pseudo-labels:
\begin{equation}
\hat{\bm{Y}}_{{{\mathcal{V}}_\text{tr}}}^{(1)},\hat{\bm{Y}}_{{{\mathcal{V}}_\text{val}}}^{(1)}={{f}_{\theta }}(\bm{X};{{\Phi }^{(0:L-1)}}({\bm{A}_{\cup }}))
\end{equation}
\begin{equation}
{{\tilde{\bm{Y}}}_{{{\mathcal{V}}_\text{val}}}}=\text{stopgrad}(\hat{\bm{Y}}_{{{\mathcal{V}}_\text{val}}}^{(1)})
\end{equation}

This gives $\frac{\partial {{\tilde{\bm{Y}}}_{{{\mathcal{V}}_{\text{val}}}}}}{\partial \theta }=0$, so validation-node predictions do not update model weights in the generative phase.

No explicit warm-up period is used in this routine. Although the pseudo-labels generated for validation nodes can be noisy at the beginning of training, they are detached from the computation graph and are not used as regression targets. Instead, they serve only as contextual inputs for the second reconstruction pass, while all learnable updates remain driven by observed training nodes with ground-truth measurements.

In the second pass, SDE samples a non-overlapping training mask ${{\mathcal{M}}_{2}}\subset {{\mathcal{V}}_{\text{tr}}}$ with ${{\mathcal{M}}_{2}}\cap {{\mathcal{M}}_{1}}=\varnothing $ and clamps validation-node features to the detached pseudo-labels. The model then predicts the new training mask:
\begin{equation}
{\bm{X}_{{{\mathcal{V}}_{\text{val}}}}}\leftarrow {{\tilde{\bm{Y}}}_{{{\mathcal{V}}_{\text{val}}}}}
\end{equation}
\begin{equation}
\hat{\bm{Y}}_{{{\mathcal{V}}_{\text{tr}}}}^{(2)}={{f}_{\theta }}(\bm{X};{{\Phi }^{(0:L-1)}}({\bm{A}_{\cup }}))
\end{equation}

The final training loss minimizes the Mean Absolute Error (MAE) exclusively over the training nodes masked in either the first or second pass:
\begin{equation}
\mathcal{L}_{\text{DRIK}} = \frac{1}{|\mathcal{M}_{1}\cup \mathcal{M}_{2}|}\sum\limits_{v\in \mathcal{M}_{1}\cup \mathcal{M}_{2}} \left| \hat{\bm{Y}}_{v}^{(\pi(v))}-\bm{Y}_{v} \right|,\quad
\pi(v)=
\begin{cases}
1, & v\in \mathcal{M}_{1},\\
2, & v\in \mathcal{M}_{2}.
\end{cases}
\end{equation}
where ${\bm{Y}_{v}}$ represents the true measurement at training node $v$. The true measurements of the validation nodes remain strictly unseen and are utilized separately only for model selection.

This design limits the effect of noisy early pseudo-labels. They are used only as detached context in the second pass, while all loss terms remain on real training-node measurements. SDE therefore improves structural consistency without supervising on validation labels.

\section{Experiments}

\begin{table}[t] 
\caption{Overview of six datasets spanning three application domains.\label{tab:dataset_overview}}
\vspace{-10pt}
\centering
\renewcommand{\arraystretch}{0.7}
\setlength{\tabcolsep}{3pt} 
\scriptsize 
\resizebox{\textwidth}{!}{
\begin{tabular}{*{9}{c}}
\toprule
\multirow{2}{*}{\textbf{Data Type}} & \multirow{2}{*}{\textbf{Dataset}} & \multicolumn{2}{c}{\textbf{Data partition}} & \multirow{2}{*}{\textbf{Region}} & \multirow{2}{*}{\textbf{Start Time}} & \multirow{2}{*}{\textbf{Samples}} & \multirow{2}{*}{\textbf{Nodes}} & \multirow{2}{*}{\textbf{Sampling Rate}} \\
\cmidrule(lr){3-4} 
& & \textbf{Temporal} & \textbf{Spatial} & & & & & \\
\midrule
\multirow{2}{*}{Traffic Speed} & METR-LA  & 7/1/2 & 3/1/1 & Los Angeles        & 3/1/2012 & 34,272  & 207 & 5 minutes \\
                               & PEMS-BAY & 7/1/2 & 3/1/1 & San Francisco      & 1/1/2017 & 52,116  & 325 & 5 minutes \\
\midrule
\multirow{2}{*}{Solar Power}   & NREL-AL  & 7/1/2 & 3/1/1 & Alabama            & 1/1/2016 & 105,120 & 137 & 5 minutes \\
                               & NREL-MD  & 7/1/2 & 3/1/1 & Maryland           & 1/1/2016 & 105,120 & 80  & 5 minutes \\
\midrule
\multirow{2}{*}{Air Quality}   & AQI-36   & 1/1/1 & 3/1/1 & Beijing            & 5/1/2014 & 8,759   & 36  & 1 hour    \\
                               & AQI      & 1/1/1 & 3/1/1 & 43 cities in China & 5/1/2014 & 8,760   & 437 & 1 hour    \\
\bottomrule
\end{tabular}
}
\vspace{-10pt}
\end{table}

\subsection{Datasets}
To evaluate the performance of DRIK, we conducted experiments on six public datasets drawn from diverse real-world scenarios, as summarized in Table~\ref{tab:dataset_overview}: two traffic datasets, two solar-power datasets, and two air-quality datasets.
(1) METR-LA \cite{ref40} contains average traffic speeds recorded by 207 detectors on Los Angeles County highways from March 1, 2012 to June 27, 2012, collected every 5 minutes. (2) PEMS-BAY \cite{ref40} comprises traffic speed measurements recorded by 325 sensors in the San Francisco Bay Area from January 1, 2017 to June 30, 2017, sampled every 5 minutes. (3) NREL-AL \cite{ref41} records solar power output from 137 photovoltaic plants in Alabama throughout 2016, with 5-minute sampling intervals. (4) NREL-MD \cite{ref41} captures solar power output from 80 photovoltaic plants in Maryland during 2016, sampled every 5 minutes. (5) AQI-36 \cite{ref42} is a subset of the Air Quality Index dataset selected from the Urban Computing Project beginning on May 1, 2014, which is commonly used in kriging studies. (6) AQI \cite{ref42} contains the full Air Quality Index dataset comprising hourly measurements of six pollutants from 437 monitoring stations across 43 Chinese cities; consistent with prior works such as GRIN \cite{ref17} and KITS \cite{ref8}, we focus specifically on PM2.5 concentrations.

\subsection{Baseline methods}
We compare our method with the following baselines. (1) Mean Imputation: Fills missing values using the average of all available node observations at each time step to avoid bias. (2) OKriging \cite{ref43}: A spatial interpolation method using Gaussian processes to model geographic correlations. (3) KNN \cite{ref44}: Estimates missing values by averaging the 10 geographically nearest neighbors. (4) KCN \cite{ref22}: Unifies GCNs and kriging using neighbor observations and attention; we evaluate its GraphSAGE variant. (5) IGNNK \cite{ref6}: Learns a transferable spatial message-passing scheme for inductive kriging via random subgraphs and signal reconstruction. (6) INCREASE \cite{ref7}: Fuses spatiotemporal signals for inductive kriging using relation-aware GRUs and multi-relation attention across heterogeneous spatial relations. (7) KITS \cite{ref8}: Bridges the train-inference gap by incorporating and supervising pseudo-labeled virtual nodes during training to improve transferability.

\subsection{Experiment settings}
For fair comparison, following prior works \cite{ref6}, we randomly select 25\% of the sensors in each dataset as unobserved locations ($\alpha$~=~0.25). The remaining observed nodes are partitioned into training (60\%), validation (20\%), and testing (20\%). Temporally, we adopt a 7:1:2 split for most datasets. To adequately capture seasonal variability in the AQI-36 and AQI datasets, we use a 1:1:1 temporal split: March, June, September, and December form the test set; February, May, August, and November constitute the validation set; and the remaining months are used for training. For data preprocessing, we apply node-specific min-max normalization for the solar datasets (NREL-AL and NREL-MD) to preserve relative generation profiles, and zero-mean normalization for the others. The spatial adjacency matrix is constructed using a thresholded Gaussian kernel. All methods are implemented using PyTorch 1.8.1 and PyTorch Lightning 1.4.0, and trained on a single NVIDIA A100 80GB GPU with random seeds fixed to 42 for reproducibility. Regarding hyperparameters, the temporal window size is set to 24, feature dimension to 64, and batch size to 32. Within the STGC module, we set m~=~1 to aggregate data from one historical, one current, and one future time interval. Models are trained using the Adam optimizer with a learning rate of $1 \times 10^{-4}$ and gradient clipping at 1.0. We train for up to 300 epochs, employing an early stopping strategy with a patience of 50 epochs. Finally, we evaluate model performance using Mean Absolute Error (MAE), Root Mean Square Error (RMSE), and Mean Absolute Percentage Error (MAPE).

To ensure reproducibility and avoid test-set tuning, all normalization statistics and model-selection decisions are determined from the training and validation splits only, and the test split is used only for final reporting. Baseline methods are trained and evaluated under the same leakage-free partitions, preprocessing pipeline, and evaluation metrics; their method-specific settings follow the original papers or released implementations whenever available. For DRIK, the main hyperparameters are fixed across datasets unless otherwise stated: the Gaussian bandwidth for distance-based graph construction is set to $\sigma=\operatorname{std}(\mathrm{dist})$, where $\mathrm{dist}$ denotes the pairwise geographical distance matrix; the SCR-reconstructed graph retains $k=5$ nearest neighbors, which provides a moderate local neighborhood for spatial aggregation; and the coordinate perturbation region is defined by the convex hull of midpoint locations between each node and its initial neighbors. The validation split is used for early stopping and checkpoint selection, while no hyperparameter is selected based on test performance.

\subsection{Experiment results and analysis}
Table~\ref{tab:drik_comparison} compares DRIK with statistical, neighborhood-based, and neural inductive kriging baselines. DRIK achieves the best result on all six datasets, while traditional statistical methods lag behind neural models because they do not model complex spatio-temporal dependencies.

\begin{table}[t]
\caption{Comparison of DRIK with existing methods on the inductive kriging task. The best results are highlighted in \textbf{bold}, while the second-best results are \underline{underlined}. $\Delta$ shows the improvement of our DRIK over the best baseline.\label{tab:drik_comparison}}
\vspace{-10pt}
\centering
\scriptsize 
\setlength{\tabcolsep}{3pt} 
\renewcommand{\arraystretch}{0.7}

\begin{tabularx}{\textwidth}{p{1.2cm} p{0.8cm} *{9}{>{\centering\arraybackslash}X}}
\toprule
\textbf{\fontsize{6pt}{7.2pt}\selectfont Dataset} & \textbf{\fontsize{6pt}{7.2pt}\selectfont Metric} & \textbf{\fontsize{6pt}{7.2pt}\selectfont Mean} & \textbf{\fontsize{6pt}{7.2pt}\selectfont OKriging} & \textbf{\fontsize{6pt}{7.2pt}\selectfont KNN} & \textbf{\fontsize{6pt}{7.2pt}\selectfont KCN} & \textbf{\fontsize{6pt}{7.2pt}\selectfont IGNNK} & \textbf{\fontsize{4.5pt}{5.4pt}\selectfont INCREASE} & \textbf{\fontsize{6pt}{7.2pt}\selectfont KITS} & \textbf{\fontsize{6pt}{7.2pt}\selectfont DRIK} & { $\mathbf{\Delta}$} \\
\midrule
\multirow{3}{*}{\begin{tabular}[c]{@{}l@{}}METR-LA\\(207)\end{tabular}} 
& MAE  & 8.272  & 7.294  & 7.987  & 7.190  & 5.801  & 5.992  & \underline{5.666}  & \textbf{5.197}  & \textbf{8.28\%}  \\
& RMSE & 11.417 & 10.277 & 12.370 & 12.470 & \underline{8.914}  & 9.198  & 8.981  & \textbf{8.101}  & \textbf{9.12\%}  \\
& MAPE & 22.133 & 18.896 & 19.820 & 23.983 & 15.581 & 16.854 & \underline{15.096} & \textbf{13.154} & \textbf{12.86\%} \\
\midrule
\multirow{3}{*}{\begin{tabular}[c]{@{}l@{}}PEMS-BAY\\(325)\end{tabular}} 
& MAE  & 4.999  & 4.874  & 5.678  & 4.676  & 3.445  & 3.599  & \underline{3.410}  & \textbf{3.218}  & \textbf{5.63\%}  \\
& RMSE & 8.474  & 8.266  & 10.431 & 9.253  & \underline{6.067}  & 6.850  & 6.445  & \textbf{5.840}  & \textbf{3.75\%}  \\
& MAPE & 12.862 & 12.412 & 14.087 & 13.514 & \underline{8.378}  & 9.457  & 8.602  & \textbf{7.728}  & \textbf{7.76\%}  \\
\midrule
\multirow{3}{*}{\begin{tabular}[c]{@{}l@{}}NREL-AL\\(137)\end{tabular}} 
& MAE  & 5.492  & 7.960  & 7.962  & 4.541  & \underline{4.531}  & 5.524  & 4.532  & \textbf{3.966}  & \textbf{12.48\%} \\
& RMSE & 8.353  & 10.580 & 10.582 & 6.697  & 6.619  & 7.950  & \underline{6.510}  & \textbf{6.357}  & \textbf{2.35\%}  \\
& MAPE & 166.221& 406.106& 410.155& 155.001& 160.523& \underline{116.402}& 177.941& \textbf{81.963} & \textbf{29.59\%} \\
\midrule
\multirow{3}{*}{\begin{tabular}[c]{@{}l@{}}NREL-MD\\(80)\end{tabular}} 
& MAE  & 11.257 & 11.947 & 11.953 & 10.961 & 11.011 & \underline{10.282} & 11.601 & \textbf{10.151} & \textbf{1.28\%}  \\
& RMSE & 16.387 & 16.455 & 16.464 & 17.032 & 17.308 & \underline{16.271} & 17.589 & \textbf{16.163} & \textbf{0.66\%}  \\
& MAPE & 294.610& 703.908& 706.322& 173.269& 195.432& \underline{147.958}& 444.394& \textbf{95.635} & \textbf{35.36\%} \\
\midrule
\multirow{3}{*}{\begin{tabular}[c]{@{}l@{}}AQI-36\\(36)\end{tabular}} 
& MAE  & 18.431 & 16.003 & \underline{14.727} & 21.963 & 20.138 & 16.963 & 19.600 & \textbf{13.443} & \textbf{8.71\%}  \\
& RMSE & 31.631 & 28.744 & \underline{26.800} & 36.647 & 33.993 & 32.854 & 34.668 & \textbf{25.550} & \textbf{4.67\%}  \\
& MAPE & 49.586 & 42.670 & \underline{37.737} & 57.988 & 69.964 & 41.619 & 76.466 & \textbf{28.433} & \textbf{24.65\%} \\
\midrule
\multirow{3}{*}{\begin{tabular}[c]{@{}l@{}}AQI\\(437)\end{tabular}} 
& MAE  & 39.718 & 23.827 & 18.376 & 21.012 & 16.315 & \underline{16.034} & 16.068 & \textbf{15.364} & \textbf{4.18\%}  \\
& RMSE & 59.968 & 39.846 & 32.490 & 35.111 & \underline{29.448} & 29.862 & 29.791 & \textbf{28.437} & \textbf{3.43\%}  \\
& MAPE & 142.226& 85.340 & 52.270 & 61.017 & 44.635 & 43.268 & \textbf{39.033} & \underline{40.180}  & -2.94\% \\
\bottomrule
\end{tabularx}
\vspace{-10pt}
\end{table}

The gains are consistent across metrics and domains. DRIK reduces MAE by 12.48\% on NREL-AL, MAPE by 24.65\% on AQI-36, and MAE/RMSE/MAPE by 8.28\%, 9.12\%, and 12.86\% on METR-LA. The improvements are not limited to one data type: traffic speed, solar power, and air quality all benefit from the proposed robust-learning design. The largest MAPE gains appear on spatially heterogeneous datasets, including NREL-MD (35.36\%), NREL-AL (29.59\%), and AQI-36 (24.65\%), indicating that the three modules are most useful when local and global spatial patterns vary strongly. This is consistent with the motivation of DRIK, because heterogeneous regions make the model more vulnerable to fixed-graph overfitting, ambiguous propagation from masked nodes, and train-inference topology mismatch.

\begin{figure}[t]
\centering
\includegraphics[width=0.72\linewidth]{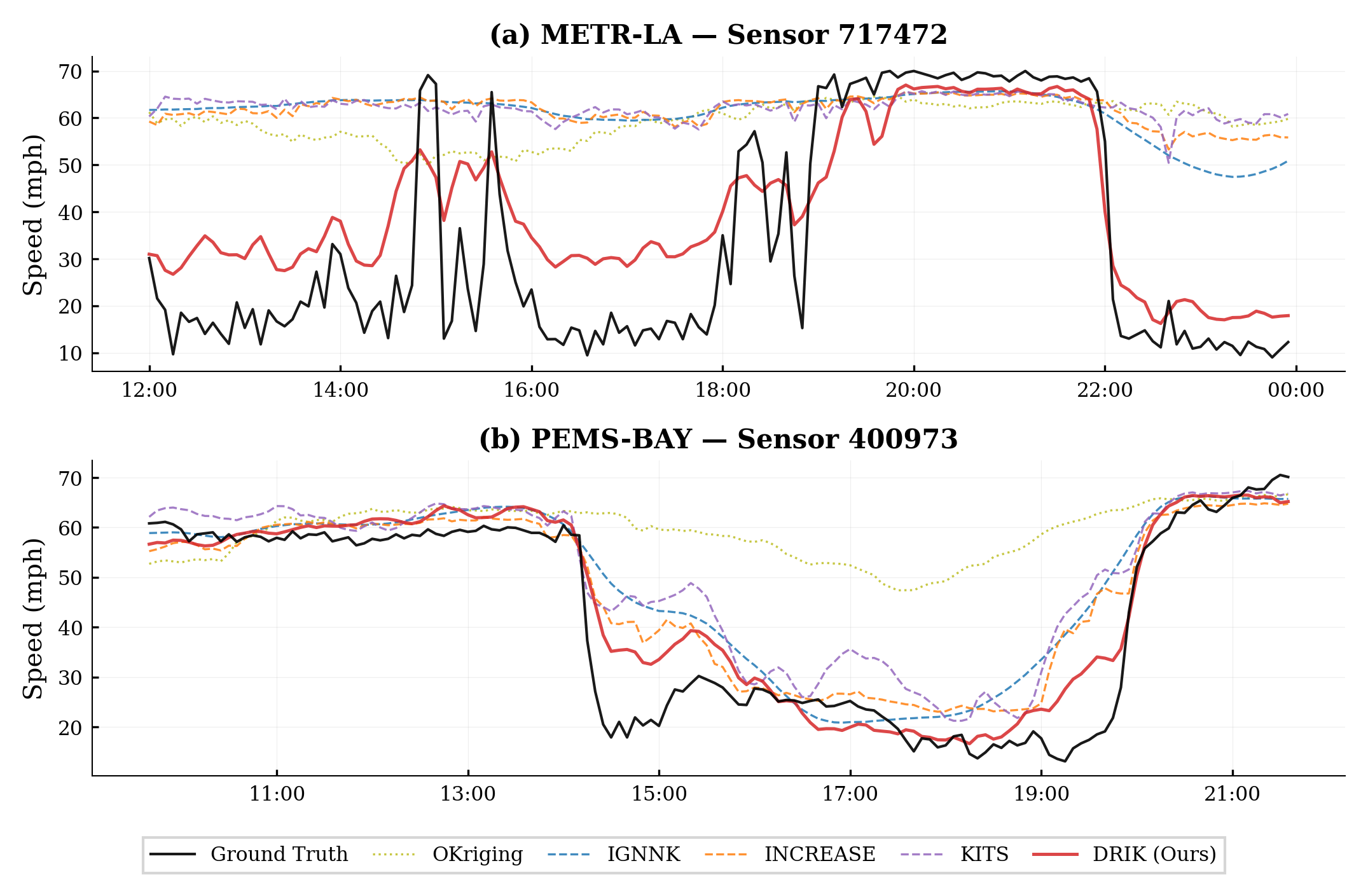}
\caption{Time series comparison of predicted traffic speed versus ground truth at two representative test sensors over a 12-hour window. (a) METR-LA, Sensor 717472, June 23, 2012, 12:00–23:55: DRIK achieves a local MAE of 11.13 mph, substantially outperforming IGNNK (29.59), INCREASE (30.30), KITS (31.00), and OKriging (29.05). (b) PEMS-BAY, Sensor 400973, June 22, 2017, 09:40–21:35: DRIK achieves a local MAE of 4.62 mph, compared with IGNNK (7.44), INCREASE (6.00), KITS (8.42), and OKriging (18.63).}
\label{fig_5}
\vspace{-10pt}
\end{figure} 

Fig.~\ref{fig_5} compares predicted and true traffic-speed series at two unseen sensors. DRIK follows sharp drops, recoveries, and regime transitions more closely than the baselines, which often over-smooth the trajectory or respond with a delay during congestion. These qualitative patterns complement the aggregate metrics: the advantage of DRIK is not only a reduction in average error, but also a better ability to preserve local temporal dynamics at nodes that were never observed during training.

\subsection{Computational cost}
We further compare the computational cost of DRIK with three representative inductive kriging baselines, namely IGNNK, INCREASE, and KITS. Since these methods rely on sampled subgraphs or sampled mini-batches, one epoch is not strictly comparable across different training routines. Therefore, we report training time per step, the number of steps required to reach the best validation checkpoint, and the wall-clock time to that checkpoint. For inference, we report both the full test time and the average time per test window.

Let $n$, $e$, and $L$ denote the number of nodes, edges, and graph-convolution layers in a sampled graph. Since DRIK uses a sparse $k$-nearest-neighbor graph with fixed $k$, the propagation cost of one forward kriging pass is linear in the number of retained edges, i.e., $\mathcal{O}(Le)$ up to feature-transformation constants. SCR adds the cost of rebuilding the perturbed $k$NN graph during training; with exact pairwise distance computation this step is $\mathcal{O}(n^{2})$ for the sampled graph, but it is not used during inference. SDE mainly increases training cost by expanding the training graph with validation-node topology and performing the two-stage pseudo-labeling routine. MFD is only a layer-wise edge-mask operation with $\mathcal{O}(e)$ cost and does not change the asymptotic order of message passing. Therefore, the additional complexity of DRIK is concentrated in training, while inference keeps the same sparse-graph forward order as the underlying kriging backbone.

\begin{table}[t]
\caption{Training and inference cost comparison on METR-LA and PEMS-BAY.\label{tab:runtime_comparison}}
\vspace{-5pt}
\centering
\renewcommand{\arraystretch}{0.75}
\scriptsize
\resizebox{\textwidth}{!}{
\begin{tabular}{llccccc}
\toprule
\textbf{Dataset} & \textbf{Method} & \textbf{Train time / step} & \textbf{Steps to best ckpt} & \textbf{Time to best ckpt} & \textbf{Full test time} & \textbf{Time / inference} \\
\midrule
\multirow{4}{*}{METR-LA}
& DRIK     & 0.258 s & 24,159 & 1.73 h & 14.38 s  & 2.10 ms \\
& IGNNK    & 0.129 s & 17,066 & 0.61 h & 14.30 s  & 2.09 ms \\
& INCREASE & 5.775 s & 2,986  & 4.79 h & 170.27 s & 24.86 ms \\
& KITS     & 0.423 s & 19,732 & 2.32 h & 19.51 s  & 2.85 ms \\
\midrule
\multirow{4}{*}{PEMS-BAY}
& DRIK     & 0.511 s & 11,412 & 1.62 h & 47.74 s  & 4.58 ms \\
& IGNNK    & 0.202 s & 26,346 & 1.48 h & 25.04 s  & 2.40 ms \\
& INCREASE & 9.645 s & 1,172  & 3.14 h & 394.30 s & 37.85 ms \\
& KITS     & 0.883 s & 24,372 & 5.98 h & 40.59 s  & 3.90 ms \\
\bottomrule
\end{tabular}
}
\vspace{-8pt}
\end{table}

As shown in Table~\ref{tab:runtime_comparison}, the two-stage SDE mechanism increases the training cost of DRIK compared with IGNNK, but the overhead remains controlled. DRIK reaches the best checkpoint faster than KITS on both METR-LA and PEMS-BAY, because KITS introduces virtual-node augmentation and cycle-consistency reconstruction. DRIK is also more efficient than INCREASE, whose node-wise temporal encoding and recurrent updates lead to much higher per-step cost.

Importantly, SDE is not executed during inference. During testing, DRIK performs only the forward kriging process with MFD, while SCR and SDE are used only during training. As a result, the inference cost remains practical: DRIK requires 2.10 ms per test window on METR-LA and 4.58 ms on PEMS-BAY. These results indicate that the proposed two-stage training design improves robustness without compromising deployment efficiency.

\subsection{Analysis of OOD properties}
\begin{figure}[t]
\centering
\includegraphics[width=\linewidth]{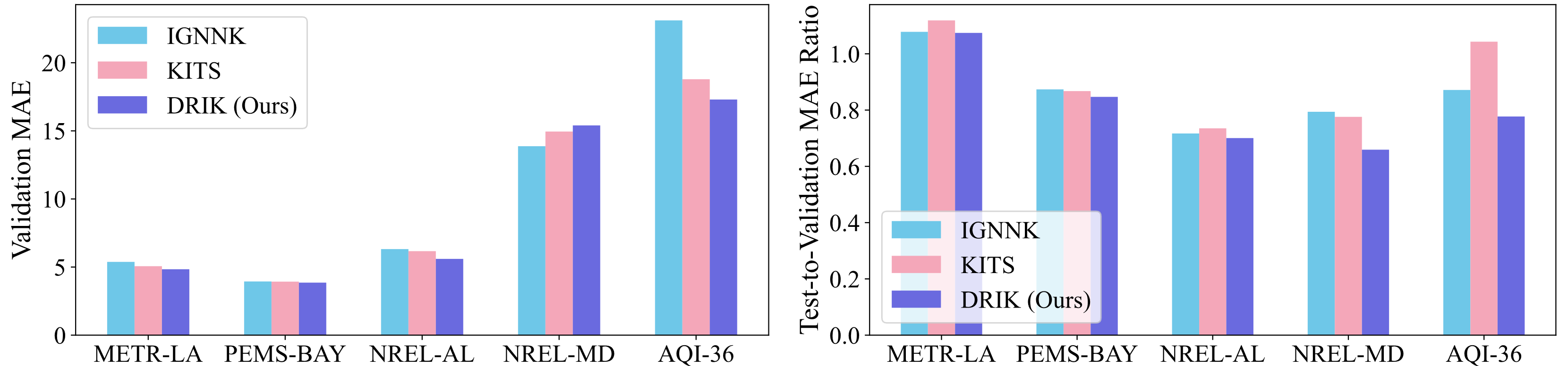}
\caption{OOD property evaluation of IGNNK, KITS, and DRIK. A smaller test-to-validation MAE ratio indicates stronger generalization ability. Note that while the validation MAE typically serves as the final evaluation metric under the conventional 2×2 partition setting, it fails to represent the model's true out-of-distribution performance due to implicit spatial information leakage.}
\label{fig_6}
\vspace{-5pt}
\end{figure}

To evaluate OOD behavior, we compare IGNNK, KITS, and DRIK using validation MAE and the test-to-validation MAE ratio (Fig.~\ref{fig_6} and Table~\ref{tab:ood_generalization}). A lower ratio indicates less degradation from validation to spatially independent testing.

\begin{table}[t] 
\caption{Quantitative results of OOD generalization analysis. MAE (Val) and MAE (Test) represent the Mean Absolute Error on the validation and test sets, respectively. The Test/Val Ratio calculates the ratio of test MAE to validation MAE, where a lower value signifies superior OOD generalization. $\Delta$ shows the improvement of DRIK over the best baseline.\label{tab:ood_generalization}}
\vspace{-5pt}
\centering
\renewcommand{\arraystretch}{0.7}
\resizebox{0.6\columnwidth}{!}{
\begin{tabular}{ll cccc}
\toprule
\textbf{Dataset} & \textbf{Metric} & \textbf{IGNNK} & \textbf{KITS} & \textbf{DRIK} & $\mathbf{\Delta}$ \\
\midrule
\multirow{3}{*}{METR-LA}  
& MAE (Val)      & 5.381  & 5.064  & 4.838  & 4.46\%  \\
& MAE (Test)     & 5.801  & 5.666  & 5.197  & 8.28\%  \\
& Test/Val Ratio & 1.078  & 1.119  & 1.074  & --      \\
\midrule
\multirow{3}{*}{PEMS-BAY} 
& MAE (Val)      & 3.944  & 3.931  & 3.768  & 4.15\%  \\
& MAE (Test)     & 3.445  & 3.410  & 3.218  & 5.63\%  \\
& Test/Val Ratio & 0.874  & 0.868  & 0.854  & --      \\
\midrule
\multirow{3}{*}{NREL-AL}  
& MAE (Val)      & 6.320  & 6.166  & 5.642  & 8.50\%  \\
& MAE (Test)     & 4.531  & 4.532  & 3.966  & 12.47\% \\
& Test/Val Ratio & 0.717  & 0.735  & 0.703  & --      \\
\midrule
\multirow{3}{*}{NREL-MD}  
& MAE (Val)      & 13.868 & 14.949 & 15.404 & -11.08\%\\
& MAE (Test)     & 11.011 & 11.601 & 10.151 & 7.81\%  \\
& Test/Val Ratio & 0.794  & 0.776  & 0.659  & --      \\
\midrule
\multirow{3}{*}{AQI-36}   
& MAE (Val)      & 23.115 & 18.787 & 17.303 & 7.90\%  \\
& MAE (Test)     & 20.138 & 19.600 & 13.443 & 31.41\% \\
& Test/Val Ratio & 0.871  & 1.043  & 0.777  & --      \\
\bottomrule
\end{tabular}
}
\vspace{-10pt}
\end{table}

The 3$\times$3 protocol shows why validation MAE alone is insufficient. Under leakage-prone splits, model selection and final evaluation may share the same spatial node domain; Table~\ref{tab:ood_generalization} shows that this can change model rankings. On NREL-MD, DRIK has higher validation MAE than IGNNK and KITS, yet obtains lower independent test MAE, indicating that fitting the validation topology does not guarantee transfer to a distinct test topology.

DRIK achieves the lowest test-to-validation MAE ratio on all evaluated datasets, and its improvement margin is generally larger on the independent test set than on the validation set. This supports the intended role of SCR, MFD, and SDE: they reduce spatial overfitting and improve robustness under strictly unseen node domains. The ratio-based comparison is also useful because absolute validation errors can vary with the difficulty of each validation split, whereas the degradation from validation to testing more directly reflects whether the learned representation transfers across spatial domains.

\begin{figure}[t]
\centering
\includegraphics[width=\linewidth]{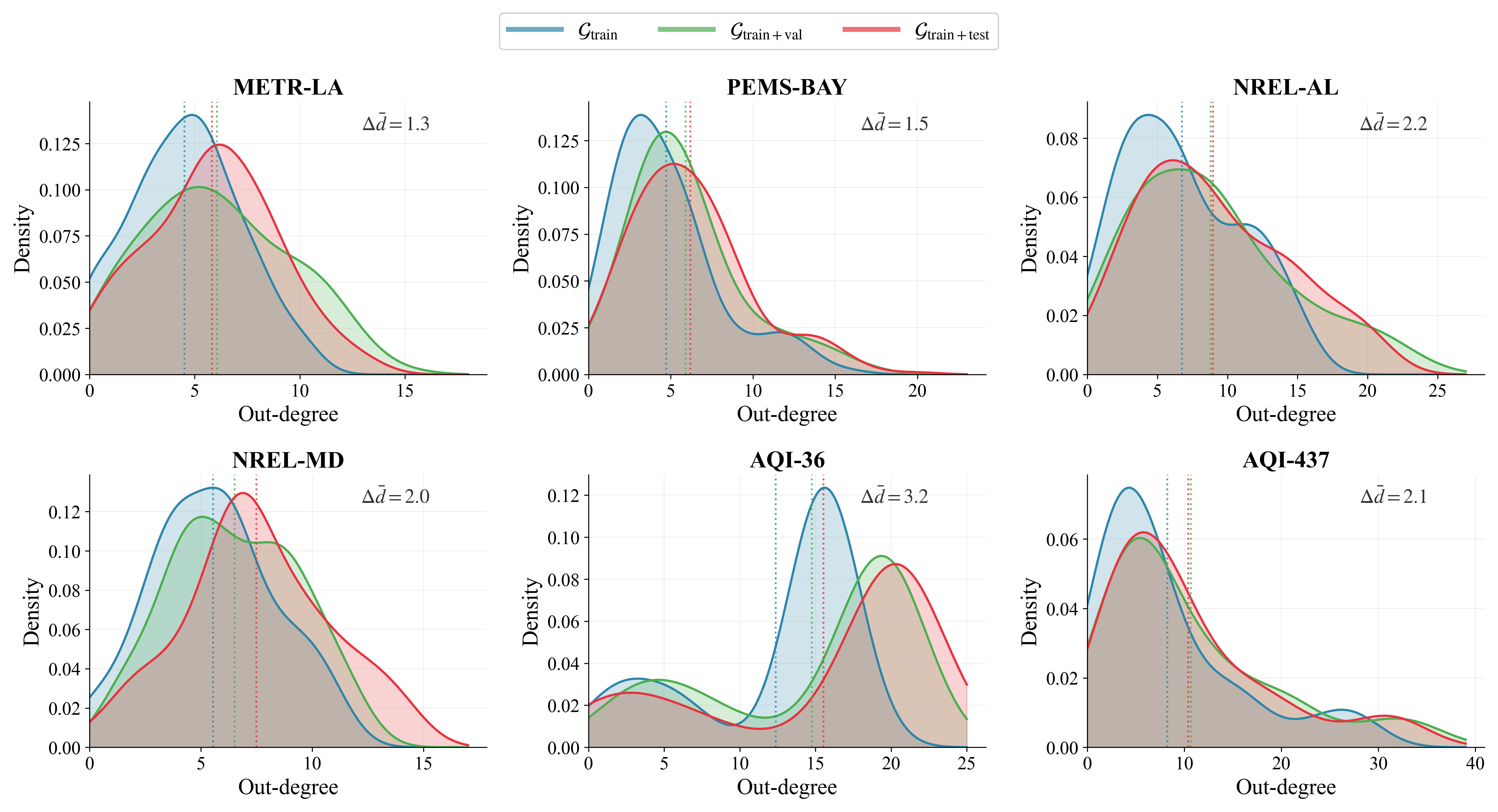}
\caption{Out-degree distribution of induced subgraphs under the 3×3 data partition across six benchmark datasets. For each dataset, three induced subgraphs are constructed: the training subgraph ${{\mathcal{G}}_\text{tr}}$ (blue), the training-plus-validation subgraph ${{\mathcal{G}}_{\text{tr}+\text{val}}}$ (green), and the training-plus-test subgraph ${{\mathcal{G}}_{\text{tr}+\text{test}}}$ (red). Dashed vertical lines mark the mean out-degree of each subgraph. $\Delta \bar{d}$ reports the absolute shift in mean out-degree between ${{\mathcal{G}}_\text{tr}}$ and ${{\mathcal{G}}_{\text{tr}+\text{test}}}$.}
\label{fig_7}
\vspace{-10pt}
\end{figure} 

We also analyze the structural shift exposed by the 3$\times$3 protocol through the out-degree distributions of ${{\mathcal{G}}_\text{tr}}$, ${{\mathcal{G}}_{\text{tr}+\text{val}}}$, and ${{\mathcal{G}}_{\text{tr}+\text{test}}}$ (Fig.~\ref{fig_7}). Since out-degree controls the receptive field of message passing, shifts in this distribution indicate different aggregation conditions at test time.

Across all datasets, ${{\mathcal{G}}_{\text{tr}+\text{test}}}$ differs from ${{\mathcal{G}}_\text{tr}}$, with mean out-degree shifts $\Delta \bar{d}$ from 1.3 on METR-LA to 3.2 on AQI-36. This confirms that the split creates a real topological gap, not only a formal sample split. SDE helps by training on ${{\mathcal{G}}_{\text{tr}+\text{val}}}$ without validation labels, which brings the training-stage topology closer to the test-stage topology. The effect is particularly clear on AQI-36, where the graph is small and each inserted node can noticeably alter local connectivity. On this dataset, where the structural shift is largest, DRIK also achieves its largest test MAE reduction of 31.41\% over the best baseline.

\subsection{Ablation study}
\subsubsection{Ablation of core modules}
Table~\ref{tab:ablation_study} demonstrates the efficacy of each proposed module within the DRIK framework. Compared with the baseline (M-0), employing SCR or MFD individually yields consistent but modest improvements, whereas using SDE alone (M-3) leads to a clear performance drop. This degradation does not indicate that validation-node topology is intrinsically harmful; rather, SDE becomes unstable when it expands the structural domain without the other two controls. Without SCR, the expanded graph remains fixed after validation nodes are inserted, so the model can overfit the static validation-augmented topology instead of learning spatially continuous representations. Without MFD, valueless or zero-padded nodes may still introduce ambiguous messages during propagation, and the enlarged graph increases the number of paths through which such unreliable information can spread. Therefore, isolated SDE increases graph density while also amplifying structural overfitting and noisy message flow, acting as a negative regularizer. When SDE is combined with SCR and MFD, SCR provides coordinate-level topological diversity and MFD suppresses ambiguous propagation from masked nodes, making the expanded validation topology useful contextual information rather than a source of noise. Accordingly, the full configuration (M-7) achieves the lowest overall error (MAE 5.197), demonstrating that the three components are complementary under the leakage-free inductive setting.

\begin{table}[t]
\caption{Component-wise ablation study of the proposed DRIK framework.\label{tab:ablation_study}}
\vspace{-5pt}
\centering
\renewcommand{\arraystretch}{0.7}
\scriptsize 
\begin{tabularx}{0.8\columnwidth}{l *{6}{>{\centering\arraybackslash}X}}
\toprule
\textbf{Model} & \textbf{SCR} & \textbf{MFD} & \textbf{SDE} & \textbf{MAE} & \textbf{RMSE} & \textbf{MAPE} \\
\midrule
M-0 &            &            &            & 6.090 & 9.372  & 16.274 \\
\midrule
M-1 & \checkmark &            &            & 5.781 & 9.115  & 16.682 \\
M-2 &            & \checkmark &            & 5.713 & 8.994  & 15.459 \\
M-3 &            &            & \checkmark & 6.419 & 10.092 & 17.056 \\
\midrule
M-4 & \checkmark & \checkmark &            & 5.368 & 8.335  & 14.685 \\
M-5 & \checkmark &            & \checkmark & 5.589 & 8.604  & 13.465 \\
M-6 &            & \checkmark & \checkmark & 5.674 & 9.031  & 15.252 \\
\midrule
M-7 & \checkmark & \checkmark & \checkmark & \textbf{5.197} & \textbf{8.101} & \textbf{13.154} \\
\bottomrule
\end{tabularx}
\end{table}

\begin{table}[t]
\caption{Performance comparison between SCR and ordinary node perturbation (NP).\label{tab:scr_np_comparison}}
\vspace{-5pt}
\centering
\renewcommand{\arraystretch}{0.7}
\scriptsize 
\setlength{\tabcolsep}{14pt} 
\begin{tabular}{llcc}
\toprule
\textbf{Dataset} & \textbf{Metric} & \textbf{SCR} & \textbf{NP} \\
\midrule
\multirow{3}{*}{METR-LA}  
& MAE  & 5.197  & 5.605 \\
& RMSE & 8.101  & 8.729 \\
& MAPE & 13.154 & 13.815 \\
\midrule
\multirow{3}{*}{PEMS-BAY} 
& MAE  & 3.218  & 3.279 \\
& RMSE & 5.840  & 5.956 \\
& MAPE & 7.728  & 7.968 \\
\midrule
\multirow{3}{*}{NREL-AL}  
& MAE  & 3.966  & 4.201 \\
& RMSE & 6.357  & 6.661 \\
& MAPE & 81.963 & 88.251 \\
\bottomrule
\end{tabular}
\vspace{-10pt}
\end{table}

\subsubsection{Analysis of spatial continuity regularization}
Table~\ref{tab:scr_np_comparison} compares SCR with ordinary node perturbation (NP). SCR consistently performs better because it perturbs coordinates and rebuilds adjacency during training, whereas NP only changes node features on a fixed graph. This distinction matters for inductive kriging: unseen nodes primarily change the spatial graph rather than the feature vector of an already known node. The comparison therefore supports the role of SCR in reducing overfitting to one discretized topology.

\subsubsection{Analysis of masked flow disambiguation}
\begin{figure}[t]
\centering
\includegraphics[width=\linewidth]{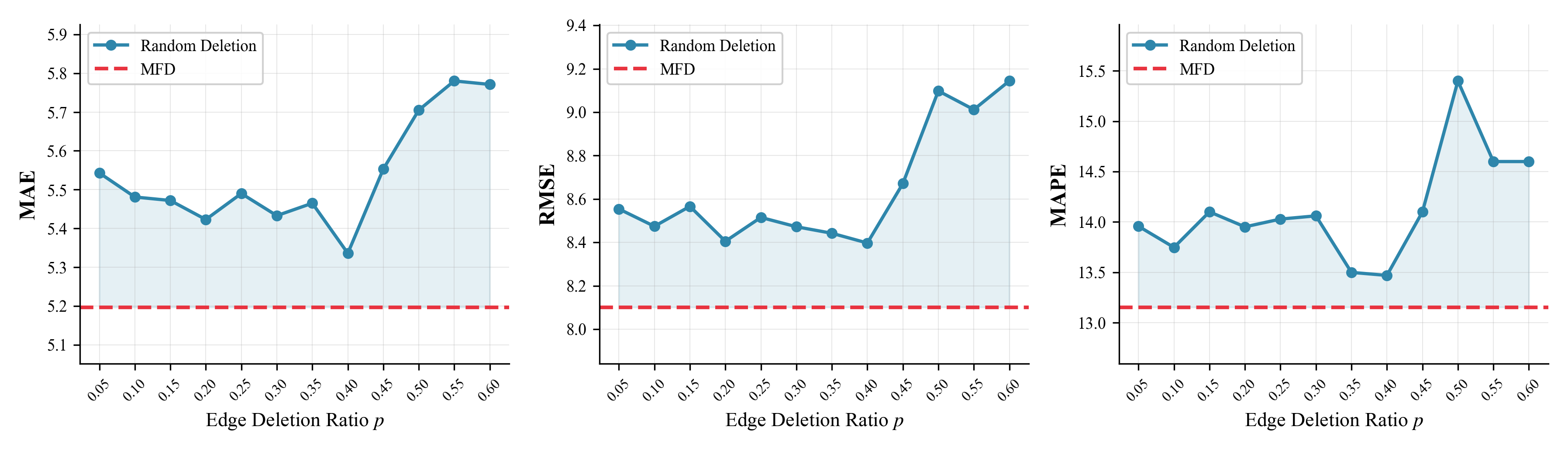}
\vspace{-27pt}
\caption{Comparison between the proposed MFD and random edge deletion on the METR-LA dataset. The horizontal dashed line denotes the performance of MFD, while the blue curve shows the results of random edge deletion under different edge deletion ratios $p$. Lower values indicate better performance. MFD consistently outperforms random deletion in terms of MAE, RMSE, and MAPE across all tested ratios.}
\label{fig_8}
\vspace{-10pt}
\end{figure}

Fig.~\ref{fig_8} compares MFD with random edge deletion under different deletion ratios $p$. MFD achieves lower MAE, RMSE, and MAPE in all tested settings. Random deletion becomes especially harmful at larger deletion ratios because it can remove informative observed-to-masked or observed-to-observed paths together with noisy ones. In contrast, MFD targets the specific ambiguity caused by zero-padded masked nodes, showing that task-aware pruning preserves useful spatial structure better than indiscriminate edge removal.

\subsubsection{Analysis of structural domain expansion}
SDE and the increment strategy in KITS both insert valueless nodes, but their inserted structures differ. KITS uses random virtual nodes, whereas DRIK uses actual validation-node topology within a decrement-increment routine. Table~\ref{tab:self_features} shows the consequence: adding self-features harms KITS, including a 20.65\% MAE increase on PEMS-BAY, but benefits DRIK, including an 18.34\% MAE reduction on METR-LA. This contrast indicates that real validation-node topology can serve as a useful spatial anchor, while random virtual nodes may introduce structural noise.

\begin{table}[t]
\caption{Impact of node self-features on the inductive kriging performance of KITS and DRIK. "w/ self" and "w/o self" denote the inclusion and exclusion of node self-features as inputs during graph convolution, respectively. $\Delta$ shows the relative improvement of the w/ self configuration over the w/o self configuration.\label{tab:self_features}}
\vspace{-5pt}
\centering
\renewcommand{\arraystretch}{0.7}
\resizebox{0.75\columnwidth}{!}{
\begin{tabular}{ll ccc ccc}
\toprule
\multirow{2.5}{*}{\textbf{Dataset}} & \multirow{2.5}{*}{\textbf{Metric}} & \multicolumn{3}{c}{\textbf{KITS}} & \multicolumn{3}{c}{\textbf{DRIK}} \\
\cmidrule(lr){3-5} \cmidrule(lr){6-8}
& & \textbf{w/o self} & \textbf{w/ self} & $\mathbf{\Delta}$ & \textbf{w/o self} & \textbf{w/ self} & $\mathbf{\Delta}$ \\
\midrule
\multirow{3}{*}{METR-LA}  
& MAE  & 5.666  & 6.211  & -9.62\%  & 6.364  & 5.197  & 18.34\% \\
& RMSE & 8.981  & 9.416  & -4.84\%  & 9.755  & 8.101  & 16.96\% \\
& MAPE & 15.096 & 16.901 & -11.96\% & 16.415 & 13.154 & 19.87\% \\
\midrule
\multirow{3}{*}{PEMS-BAY} 
& MAE  & 3.410  & 4.114  & -20.65\% & 3.669  & 3.218  & 12.29\% \\
& RMSE & 6.445  & 7.913  & -22.78\% & 6.414  & 5.840  & 8.95\% \\
& MAPE & 8.602  & 11.590 & -34.74\% & 9.282  & 7.728  & 16.74\% \\
\bottomrule
\end{tabular}
}
\vspace{-10pt}
\end{table}

\subsection{Sensitivity Analysis}
\subsubsection{Sensitivity to Spatial Sampling Variations}
\begin{figure}[t]
\centering
\includegraphics[width=0.9\linewidth]{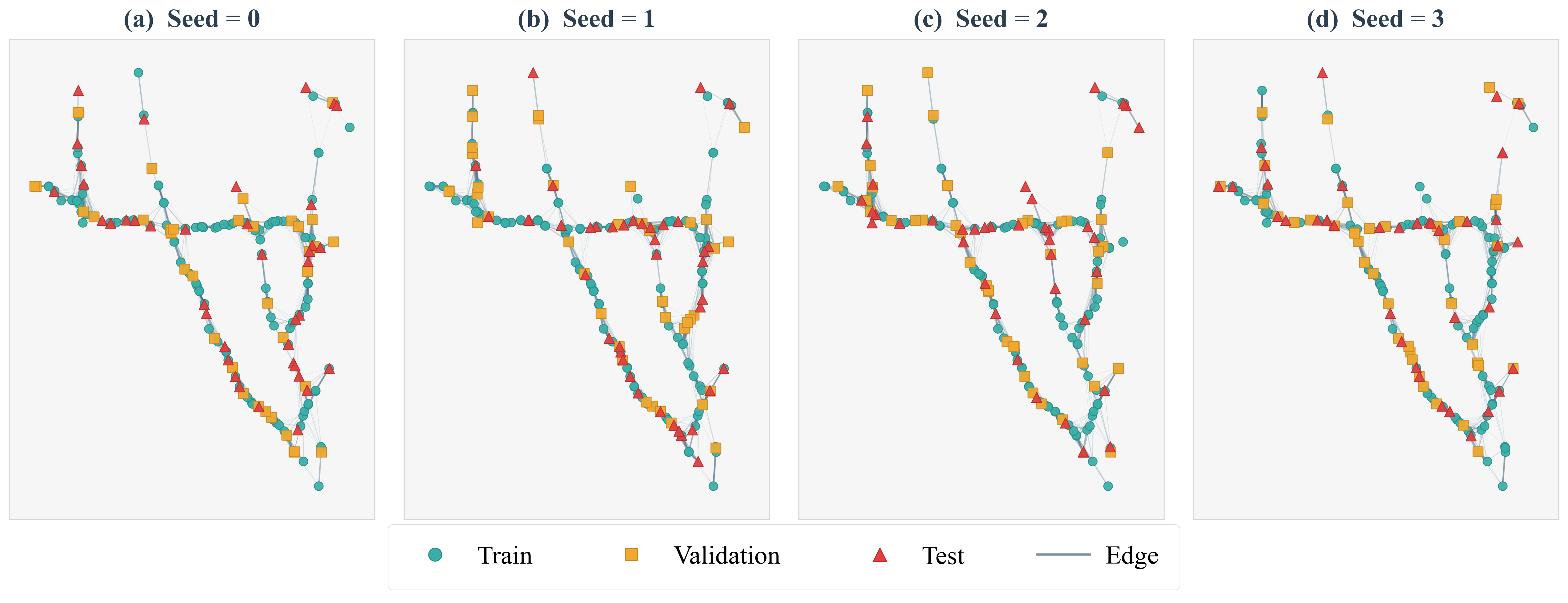}
\vspace{-10pt}
\caption{Examples of missing-pattern generation on the METR-LA dataset with a missing ratio of $\alpha$ = 25\%.}
\label{fig_9}
\end{figure}

We evaluate robustness to spatial sampling variation by changing which nodes are observed or unobserved while keeping the missing ratio fixed. Fig.~\ref{fig_9} shows that, at $\alpha$ = 25\% on METR-LA, different random seeds produce distinct missing patterns and local structures. This setting tests whether a method remains stable across multiple spatial partitions rather than a single fixed split.

\begin{figure}[t]
\centering
\includegraphics[width=0.9\linewidth]{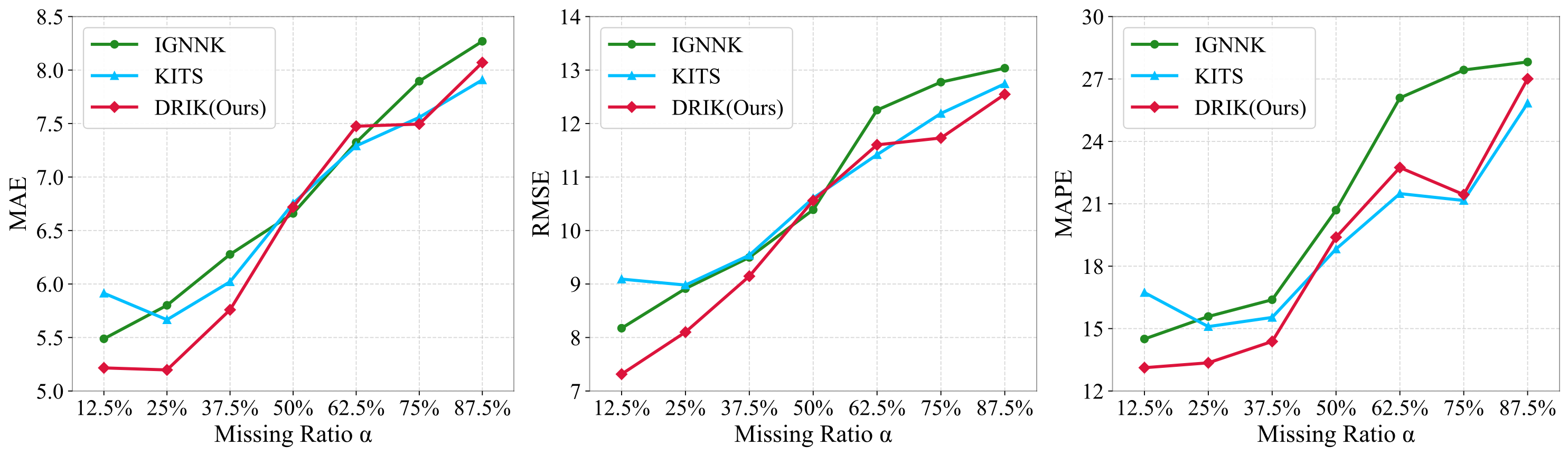}
\vspace{-8pt}
\caption{Comparisons with different missing ratios $\alpha$.}
\label{fig_10}
\vspace{-10pt}
\end{figure} 

Table~\ref{tab:seed_robustness} shows that DRIK obtains the best MAE under all four seeds, with an average of 5.753±0.603 versus 6.292±0.475 for KITS. Seed 2 is harder for all methods, which indicates that some missing-node patterns produce less favorable spatial coverage than others. DRIK still ranks first under this difficult split and improves over the best baseline by 1.710\%, suggesting that its gains are not tied to a single favorable random partition.

\begin{table}[t]
\caption{Performance of inductive methods under four different random seeds on the METR-LA dataset. The best results are shown in \textbf{bold}, and the second-best results are \underline{underlined}. $\Delta$ shows the improvement of DRIK over the best baseline.\label{tab:seed_robustness}}
\vspace{-5pt}
\centering
\renewcommand{\arraystretch}{0.7}
\resizebox{0.6\columnwidth}{!}{
\begin{tabular}{l ccccc}
\toprule
\textbf{Method} & \textbf{Seed 0} & \textbf{Seed 1} & \textbf{Seed 2} & \textbf{Seed 3} & \textbf{Mean $\pm$ Std} \\
\midrule
Mean       & 8.272 & 9.318 & 8.282 & 8.932 & 8.701 $\pm$ 0.473 \\
OKriging   & 7.294 & 7.907 & 7.793 & 7.663 & 7.664 $\pm$ 0.239 \\
KNN        & 7.987 & 8.332 & 8.881 & 8.610 & 8.453 $\pm$ 0.374 \\
KCN        & 7.190 & 7.281 & 8.112 & 7.101 & 7.421 $\pm$ 0.455 \\
IGNNK      & 5.801 & \underline{6.006} & 7.579 & 6.479 & 6.466 $\pm$ 0.726 \\
INCREASE   & 5.992 & 6.221 & 7.680 & 6.978 & 6.718 $\pm$ 0.732 \\
KITS       & \underline{5.666} & 6.031 & \underline{6.848} & \underline{6.621} & \underline{6.292 $\pm$ 0.475} \\
DRIK(Ours) & \textbf{5.197} & \textbf{5.450} & \textbf{6.731} & \textbf{5.635} & \textbf{5.753 $\pm$ 0.603} \\
\midrule 
$\Delta$   & \textbf{8.280\%} & \textbf{9.260\%} & \textbf{1.710\%} & \textbf{14.890\%} & \textbf{8.570\%} \\
\bottomrule
\end{tabular}
}
\vspace{-10pt}
\end{table}

\subsubsection{Sensitivity to Data Sparsity}
Fig.~\ref{fig_10} evaluates missing ratios from 12.5\% to 87.5\%. DRIK is strongest at low and medium sparsity, with MAE values of 5.216, 5.197, and 5.759 at 12.5\%, 25\%, and 37.5\%, corresponding to 4.96\%, 8.28\%, and 4.35\% improvements over the best baseline. These ranges are also common in practical sensing systems, where a portion of locations is uninstrumented but the observed network still provides sufficient spatial support. Under extreme sparsity, its margin narrows but remains competitive, suggesting that SDE partly offsets the node-isolation risk introduced by MFD.

\section{Limitations}
Although DRIK improves leakage-free inductive kriging, several limitations remain. First, the spatial graph is constructed from coordinates using a Gaussian kernel and a $k$-nearest-neighbor rule. This choice is broadly applicable when detailed physical topology is unavailable, but it may be insufficient when road connectivity, travel time, directionality, terrain, or other constraints dominate spatial dependence. Future work can incorporate physical-network distances, anisotropic kernels, or physics-informed graphs.

Second, SDE uses validation-node topology without labels to reduce the train-inference structural gap, but it increases training cost. It also assumes that the unlabeled validation topology is reasonably representative of the structural conditions encountered during testing. When validation nodes are sparse or topologically biased, this structural context may be incomplete and should be used with caution.

Third, our experiments cover six public datasets from traffic, solar power, and air quality domains. Broader deployment scenarios may involve stronger non-stationarity, sensor failures, irregular sampling, or domain-specific measurement noise. Future work should evaluate DRIK under such settings and further examine uncertainty across more spatial partitions and real deployment constraints.

\section{Conclusions}
This work identifies implicit spatial leakage in conventional inductive kriging evaluation and proposes a leakage-free 3$\times$3 partition that separates training, validation, and testing in both space and time. On this basis, DRIK improves structural robustness through SCR, MFD, and SDE, respectively addressing coordinate discretization, ambiguous masked-node propagation, and train-inference topology mismatch. Experiments on six spatio-temporal datasets show that DRIK reduces MAE by up to 12.48\%, improves OOD generalization, and remains competitive under high missing ratios. The results also show that leakage-free evaluation can reveal differences that are hidden by validation-only or spatially reused protocols. More broadly, the study suggests that inductive kriging should be assessed not only by interpolation accuracy on held-out values, but also by whether the model remains reliable when the spatial support itself changes. Future work will extend this framework to broader continuous monitoring systems and investigate topology-constrained perturbations when detailed physical network information is available.

\section*{Acknowledgments}
The authors acknowledge support from the National Key Research and Development Program of China (Grant No.~2024YFB2605600) and the National Natural Science Foundation of China (Grant No.~52121005).


\begin{thebibliography}{00}

\bibitem{ref1}
G. Zou, Z. Zhou, R. Weibel, Y. Li, T. Wang, Z. Liu, W. Ding, and C. Fu. Multi-graph spatio-temporal network for traffic accident risk forecasting, \textit{Pattern Recognit.}, p. 112784, 2025.

\bibitem{ref2}
C. Yu, F. Wang, Y. Wang, Z. Shao, T. Sun, D. Yao, and Y. Xu, MGSFformer: A Multi-Granularity Spatiotemporal Fusion Transformer for air quality prediction, \textit{Inf. Fusion}, vol. 113, p. 102607, 2025.

\bibitem{ref3}
J. Ye, C. Zhou, X. Zhou, Y. Huang, and C. Zhao. CVACL-MA: Comprehensive variate analysis and collaborative learning with multi-adapter for multivariate time series forecasting, \textit{Pattern Recognit.}, vol. 169, p. 111936, 2026.

\bibitem{ref4}
Y. Liang, K. Ouyang, L. Jing, S. Ruan, Y. Liu, J. Zhang, D. S. Rosenblum, and Y. Zheng, UrbanFM: Inferring Fine-Grained Urban Flows, in \textit{Proc. 25th ACM SIGKDD Int. Conf. Knowl. Discov. Data Min.}, 2019, pp. 3132--3142.

\bibitem{ref5}
T. Seo, A. M. Bayen, T. Kusakabe, and Y. Asakura, Traffic state estimation on highway: A comprehensive survey, \textit{Annu. Rev. Control}, vol. 43, pp. 128--151, 2017.

\bibitem{ref6}
Y. Wu, D. Zhuang, A. Labbe, and L. Sun, Inductive Graph Neural Networks for Spatiotemporal Kriging, in \textit{Proc. AAAI Conf. Artif. Intell.}, vol. 35, 2021, pp. 4478--4485.

\bibitem{ref7}
C. Zheng, X. Fan, C. Wang, J. Qi, C. Chen, and L. Chen, INCREASE: Inductive Graph Representation Learning for Spatio-Temporal Kriging, in \textit{Proc. ACM Web Conf. 2023}, 2023, pp. 673--683.

\bibitem{ref8}
Q. Xu, C. Long, Z. Li, S. Ruan, R. Zhao, and Z. Li, KITS: Inductive Spatio-Temporal Kriging with Increment Training Strategy, in \textit{Proc. AAAI Conf. Artif. Intell.}, vol. 39, 2025, pp. 12945--12953.

\bibitem{ref10}
W. Zhu, X. Zhou, S. Lan, W. Wang, Z. Hou, Y. Ren, and T. Pan, A dual branch graph neural network based spatial interpolation method for traffic data inference in unobserved locations, \textit{Inf. Fusion}, vol. 114, p. 102703, 2025.

\bibitem{ref11}
H. Yu, W. Liu, Y. Wang, B. Liu, D. Tao, D. Zeng, and H. Chen. Joint subgraph independence for graph out-of-distribution generalization, \textit{Pattern Recognit.}, p. 112551, 2025.

\bibitem{ref12}
H. Li, X. Wang, Z. Zhang, and W. Zhu, Out-of-Distribution Generalization on Graphs: A Survey, \textit{IEEE Trans. Pattern Anal. Mach. Intell.}, vol. 47, pp. 10490--10512, 2025.

\bibitem{ref13}
T. Zhou, H. Shan, A. Banerjee, and G. Sapiro, Kernelized Probabilistic Matrix Factorization: Exploiting Graphs and Side Information, in \textit{Proc. 2012 SIAM Int. Conf. Data Min. SDM}, Society for Industrial and Applied Mathematics, 2012, pp. 403--414.

\bibitem{ref14}
K. Takeuchi, H. Kashima, and N. Ueda, Autoregressive Tensor Factorization for Spatio-Temporal Predictions, in \textit{2017 IEEE Int. Conf. Data Min. ICDM}, 2017, pp. 1105--1110.

\bibitem{ref15}
L. Deng, X.-Y. Liu, H. Zheng, X. Feng, and Y. Chen, Graph Spectral Regularized Tensor Completion for Traffic Data Imputation, \textit{IEEE Trans. Intell. Transp. Syst.}, vol. 23, pp. 10996--11010, 2022.

\bibitem{ref16}
M. T. Bahadori, Q. Yu, and Y. Liu, Fast Multivariate Spatio-temporal Analysis via Low Rank Tensor Learning, \textit{Adv. Neural Inf. Process. Syst.}, vol. 27, 2014.

\bibitem{ref17}
A. Cini, I. Marisca, and C. Alippi, Filling the Gaps: Multivariate Time Series Imputation by Graph Neural Networks, in \textit{Int. Conf. Learn. Represent.}, pp. 1--20, 2022.

\bibitem{ref18}
Z. Zhang, J. Yang, X. Chen, and Y. Wu. Graph neural networks with flow conservation constraints for real-time origin--destination matrix completion, \textit{Pattern Recognit.}, vol. 170, p. 112046, 2026.

\bibitem{ref45}
C. Yu, F. Wang, Z. Shao, T. Qian, Z. Zhang, W. Wei, Z. An, Q. Wang, and Y. Xu, GinAR+: A Robust End-to-End Framework for Multivariate Time Series Forecasting With Missing Values, \textit{IEEE Trans. Knowl. Data Eng.}, vol. 37, no. 8, pp. 4635--4648, 2025.

\bibitem{ref20}
M. Liu, H. Huang, H. Feng, L. Sun, B. Du, and Y. Fu, PriSTI: A Conditional Diffusion Framework for Spatiotemporal Imputation, in \textit{2023 IEEE 39th Int. Conf. Data Eng. ICDE}, 2023, pp. 1927--1939.

\bibitem{ref21}
M. Jin, H. Y. Koh, Q. Wen, D. Zambon, C. Alippi, G. I. Webb, I. King, and S. Pan, A Survey on Graph Neural Networks for Time Series: Forecasting, Classification, Imputation, and Anomaly Detection, \textit{IEEE Trans. Pattern Anal. Mach. Intell.}, vol. 46, pp. 10466--10485, 2024.

\bibitem{ref22}
G. Appleby, L. Liu, and L.-P. Liu, Kriging Convolutional Networks, in \textit{Proc. AAAI Conf. Artif. Intell.}, vol. 34, 2020, pp. 3187--3194.

\bibitem{ref23}
Y. Wu, D. Zhuang, M. Lei, A. Labb\'e, and L. Sun, Spatial Aggregation and Temporal Convolution Networks for Real-Time Kriging, \textit{arXiv:2109.12144}, 2021.

\bibitem{ref24}
J. Hu, Y. Liang, Z. Fan, L. Liu, Y. Yin, and R. Zimmermann, Decoupling Long- and Short-Term Patterns in Spatiotemporal Inference, \textit{IEEE Trans. Neural Netw. Learn. Syst.}, vol. 35, pp. 16328--16340, 2024.

\bibitem{ref25}
T. Wei, Y. Lin, S. Guo, Y. Lin, Y. Zhao, X. Jin, Z. Wu, and H. Wan, Inductive and adaptive graph convolution networks equipped with constraint task for spatial--temporal traffic data kriging, \textit{Knowl.-Based Syst.}, vol. 284, p. 111325, 2024.

\bibitem{ref26}
S. Gui, X. Li, L. Wang, and S. Ji, GOOD: A Graph Out-of-Distribution Benchmark, \textit{Adv. Neural Inf. Process. Syst.}, vol. 35, pp. 2059--2073, 2022.

\bibitem{ref46}
F. Cheng and H. Liu, Charging strategies optimization for lithium-ion battery: Heterogeneous ensemble surrogate model-assisted advanced multi-objective optimization algorithm, \textit{Energy Convers. Manag.}, vol. 342, p. 120170, 2025.

\bibitem{ref47}
F. Cheng, H. Liu, and X. Lv, MetaGNSDformer: Meta-learning enhanced gated non-stationary informer with frequency-aware attention for point-interval remaining useful life prediction of lithium-ion batteries, \textit{Adv. Eng. Inform.}, vol. 69, p. 103798, 2026.

\bibitem{ref27}
J. Ma, P. Cui, K. Kuang, X. Wang, and W. Zhu, Disentangled Graph Convolutional Networks, in \textit{Proc. 36th Int. Conf. Mach. Learn.}, 2019, pp. 4212--4221.

\bibitem{ref28}
Y. Yang, Z. Feng, M. Song, and X. Wang, Factorizable Graph Convolutional Networks, \textit{Adv. Neural Inf. Process. Syst.}, vol. 33, pp. 20286--20296, 2020.

\bibitem{ref29}
H. Li, X. Wang, Z. Zhang, and W. Zhu, OOD-GNN: Out-of-Distribution Generalized Graph Neural Network, \textit{IEEE Trans. Knowl. Data Eng.}, vol. 35, pp. 7328--7340, 2023.

\bibitem{ref30}
R. Chen, H. Zhu, B. Xiao, and T. Ma. CAM: Causality-driven Adaptive Sparsity and Hierarchical Memory for robust out-of-distribution learning in GNNs, \textit{Pattern Recognit.}, vol. 168, p. 111812, 2025.

\bibitem{ref31}
Y. Wu, X. Wang, A. Zhang, X. He, and T.-S. Chua, Discovering Invariant Rationales for Graph Neural Networks, in \textit{Int. Conf. Learn. Represent.}, pp. 1--22, 2022.

\bibitem{ref32}
Y. Wu, A. Bojchevski, and H. Huang, Adversarial Weight Perturbation Improves Generalization in Graph Neural Networks, in \textit{Proc. AAAI Conf. Artif. Intell.}, vol. 37, 2023, pp. 10417--10425.

\bibitem{ref33}
W. Feng, J. Zhang, Y. Dong, Y. Han, H. Luan, Q. Xu, Q. Yang, E. Kharlamov, and J. Tang, Graph Random Neural Networks for Semi-Supervised Learning on Graphs, \textit{Adv. Neural Inf. Process. Syst.}, vol. 33, pp. 22092--22103, 2020.

\bibitem{ref34}
Z. Yin, Z. Wang, H. Liu, C. Li, M. Yao, S. Li, F. Li, J. Ren, and Y. Yang. PUA: Pseudo-Features Made Useful Again for Robust Graph Node Classification under Distribution Shift, \textit{Pattern Recognit.}, p. 113185, 2026.

\bibitem{ref35}
T. Zhao, Y. Liu, L. Neves, O. Woodford, M. Jiang, and N. Shah, Data Augmentation for Graph Neural Networks, in \textit{Proc. AAAI Conf. Artif. Intell.}, vol. 35, 2021, pp. 11015--11023.

\bibitem{ref36}
H. Park, S. Lee, S. Kim, J. Park, J. Jeong, K.-M. Kim, J.-W. Ha, and H. J. Kim, Metropolis-Hastings Data Augmentation for Graph Neural Networks, \textit{Adv. Neural Inf. Process. Syst.}, vol. 34, pp. 19010--19020, 2021.

\bibitem{ref37}
Y. You, T. Chen, Y. Sui, T. Chen, Z. Wang, and Y. Shen, Graph Contrastive Learning with Augmentations, \textit{Adv. Neural Inf. Process. Syst.}, vol. 33, pp. 5812--5823, 2020.

\bibitem{ref38}
G. Liu, T. Zhao, J. Xu, T. Luo, and M. Jiang, Graph Rationalization with Environment-based Augmentations, in \textit{Proc. 28th ACM SIGKDD Conf. Knowl. Discov. Data Min.}, 2022, pp. 1069--1078.

\bibitem{ref39}
B. E. Boser, I. M. Guyon, and V. N. Vapnik, A training algorithm for optimal margin classifiers, in \textit{Proc. Fifth Annu. Workshop Comput. Learn. Theory}, Association for Computing Machinery, New York, NY, USA, 1992, pp. 144--152.

\bibitem{ref40}
Y. Li, R. Yu, C. Shahabi, and Y. Liu, Diffusion Convolutional Recurrent Neural Network: Data-Driven Traffic Forecasting, in \textit{Int. Conf. Learn. Represent.}, pp. 1--16, 2018.

\bibitem{ref41}
A. Bloom, A. Townsend, D. Palchak, J. Novacheck, J. King, C. Barrows, E. Ibanez, M. O’Connell, G. Jordan, B. Roberts, C. Draxl, and K. Gruchalla, \textit{Eastern Renewable Generation Integration Study}, 2016.

\bibitem{ref42}
X. Yi, Y. Zheng, J. Zhang, and T. Li, ST-MVL: Filling Missing Values in Geo-Sensory Time Series Data, in \textit{Proc. 25th Int. Joint Conf. Artif. Intell.}, 2016, pp. 1--7.

\bibitem{ref43}
N. Cressie and C. K. Wikle, \textit{Statistics for Spatio-Temporal Data}, John Wiley \& Sons, 2011.

\bibitem{ref44}
T. Cover and P. Hart, Nearest neighbor pattern classification, \textit{IEEE Trans. Inf. Theory}, vol. 13, pp. 21--27, 1967.

\end{thebibliography}
\end{document}